%% file: main.tex
\documentclass{article}

\usepackage{PRIMEarxiv}
\usepackage{enumitem}

\usepackage[utf8]{inputenc} 
\usepackage[T1]{fontenc}    
\usepackage{hyperref}       
\usepackage{url}            
\usepackage{booktabs}       
\usepackage{amsfonts}       
\usepackage{nicefrac}       
\usepackage{microtype}      
\usepackage{lipsum}
\usepackage{fancyhdr}       
\usepackage{graphicx}       
\usepackage{amsmath}
\usepackage{amssymb}
\usepackage{diagbox}
\usepackage{float}
 
\DeclareMathOperator*{\argmax}{argmax} 
\input{algorithm}
\graphicspath{{media/}}     
\newcommand{\comment}[1]{}
\newcommand{\A}{\text{A}}
\newcommand{\C}{\text{C}}
\newcommand{\G}{\text{G}}
\newcommand{\T}{\text{T}}
\title{Marginalized Beam Search Algorithms for Hierarchical HMMs
}

\author{
  Xuechun Xu, Joakim Jaldén\\
  EECS\\
  KTH, Loyal Institution of Technology\\
  Stockholm, Sweden\\
  \texttt{chunxxc@gmail.com, jalden@kth.se} \\
}

\begin{document}
\maketitle

\begin{abstract}
Inferring a state sequence from a sequence of measurements is a fundamental problem in bioinformatics and natural language processing. The Viterbi and the Beam Search (BS) algorithms are popular inference methods, but they have limitations when applied to Hierarchical Hidden Markov Models (HHMMs), where the interest lies in the outer state sequence. The Viterbi algorithm can not infer outer states without inner states, while the BS algorithm requires marginalization over prohibitively large state spaces. We propose two new algorithms to overcome these limitations: the greedy marginalized BS algorithm and the local focus BS algorithm. We show that they approximate the most likely outer state sequence with higher performance than the Viterbi algorithm, and we evaluate the performance of these algorithms on an explicit duration HMM with simulation and nanopore base calling data.
\end{abstract}


\input{body}
\input{results}
\section*{Acknowledgments}

The authors acknowledge support from Patrik Ståhl and Nayanika Bhalla, and the National Genomics Infrastructure (NGI) in Stockholm. This work has been supported by the Swedish Research Council Research Environment Grant QuantumSense [VR 2018-06169].

\newpage

\input{MGBS_algo}
\newpage

\input{LFBS_algo}

\newpage
\bibliographystyle{unsrt}  
\bibliography{references}
\newpage
\input{appendix}

\end{document}

%% file: algorithm.tex
\usepackage{algorithm}
\usepackage{algpseudocode}

\errorcontextlines\maxdimen

\makeatletter
    \newcommand*{\algrule}[1][\algorithmicindent]{\makebox[#1][l]{\hspace*{.5em}\thealgruleextra\vrule height \thealgruleheight depth \thealgruledepth}}%
\newcommand*{\thealgruleextra}{}
\newcommand*{\thealgruleheight}{.75\baselineskip}
\newcommand*{\thealgruledepth}{.25\baselineskip}

\newcount\ALG@printindent@tempcnta
\def\ALG@printindent{%
    \ifnum \theALG@nested>0
        \ifx\ALG@text\ALG@x@notext
        \else
            \unskip
            \addvspace{-1pt}
            \ALG@printindent@tempcnta=1
            \loop
                \algrule[\csname ALG@ind@\the\ALG@printindent@tempcnta\endcsname]%
                \advance \ALG@printindent@tempcnta 1
            \ifnum \ALG@printindent@tempcnta<\numexpr\theALG@nested+1\relax
            \repeat
        \fi
    \fi
    }%
\usepackage{etoolbox}
\patchcmd{\ALG@doentity}{\noindent\hskip\ALG@tlm}{\ALG@printindent}{}{\errmessage{failed to patch}}
\makeatother

\newbox\statebox
\newcommand{\myState}[1]{%
    \setbox\statebox=\vbox{#1}%
    \edef\thealgruleheight{\dimexpr \the\ht\statebox+1pt\relax}%
    \edef\thealgruledepth{\dimexpr \the\dp\statebox+1pt\relax}%
    \ifdim\thealgruleheight<.75\baselineskip
        \def\thealgruleheight{\dimexpr .75\baselineskip+1pt\relax}%
    \fi
    \ifdim\thealgruledepth<.25\baselineskip
        \def\thealgruledepth{\dimexpr .25\baselineskip+1pt\relax}%
    \fi
    \State #1%
    \def\thealgruleheight{\dimexpr .75\baselineskip+1pt\relax}%
    \def\thealgruledepth{\dimexpr .25\baselineskip+1pt\relax}%
}

%% file: body.tex
\section{Introduction}
Many research fields require modelling the dependencies between a sequence of states $\hat{S}_{1:L} \triangleq \{\hat{S}_l\}_{l=1}^L$ and a sequence of observations $X_{1:T} \triangleq \{X_t\}_{t=1}^T$, where the state sequence length $L$ is smaller than the observation sequence length $T$. Usually, the length difference implies that a variable number of consecutive observations depend on or are dominated by any one state $\hat{S}_l$. Example state and observation sequence pairs are words and speech signals in speech recognition \cite{Kamaric1999, speech,HHMMhealthcare,IHHMM,CVDHMM}, target activities and video frames in behavioural modelling \cite{HHMManimal,HHMMmovement,HHMMdementia,HHMMonline}, or nucleotides and ion current measurements in nanopore base calling \cite{Jain2016, Lokatt,HMMRATAC}.

Hierarchical hidden Markov models (HHMMs) \cite{HHMM} have emerged as a common tool for modelling such scenarios. Referring to $\hat{S}_l$ as the sequence-aligned state, the HHMM is built upon a hierarchy of time-aligned (observation-aligned) states $S_t^r$, where $r=1,2,...,R$ is the level (rank) in the hierarchy. Each level represents a specific temporal scale or abstraction level of the observations. We refer to the original state $\hat{S}_l$ and the highest level state $S^1_t$ as the sequence- and time-aligned \textit{outer state}, respectively, and require that $\hat{S}_l$ and $S^1_t$ belong to the same state space $\mathbb{S} = \mathbb{S}^1$. The lower level states $S^r_t$ for $r=2,\ldots,R$ are referred to as the \textit{inner states} and belong to some, possibly different, individual state spaces $\mathbb{S}^r$. For convenience, we will use $S_t \triangleq (S_t^1,\ldots,S_t^R)$ to denote the joint set of time-aligned states of all levels at time $t$. Under some additional constraints to be defined later, one can introduce a many-to-one (surjective) mapping from $S_{1:T} \triangleq \{S_t\}_{t=1}^T$ to $\hat{S}_{1:L}$ such that $\hat{S}_{l_t} = S^1_t$ for some sequence $\{l_t\}_{t=1}^T$ with $1\leq l_t < l_{t+1}\leq L$. 
Essentially, the hierarchy of states allows the outer states to be aligned with the observations. 
With the time-aligned states, the Viterbi algorithm \cite{LTHHMM} provides a linear-time solution to
\begin{equation} \label{eq:inference}
\argmax_{S_{1:T}} P(S_{1:T} | X_{1:T}) = \argmax_{S_{1:T}} P(S_{1:T}, X_{1:T}) \, ,
\end{equation}
i.e., a way to infer the most likely state sequence over all levels of the hierarchy. 
However, inferring the most likely set of \emph{sequence-aligned} outer states by solving
\begin{equation} \label{eq:decoding}
\argmax_{\hat{S}_{1:L}} P(\hat{S}_{1:L} | X_{1:T}) = \argmax_{\hat{S}_{1:L}} P(\hat{S}_{1:L},X_{1:T})
\end{equation}
typically remains intractable due to the need for marginalizing over the set of $S_{1:T}$ that map to any $\hat{S}_{1:L}$. The problem in \eqref{eq:decoding}, which we will refer to simply as the \emph{decoding} problem, is the key inference problem studied in this work.

Numerous studies have been dedicated to finding approximate solutions to the decoding problem since $\hat{S}_{1:L}$ is the main quantity of interest in applications such as speech recognition or nanopore sequencing. The alignment information in these applications is typically of secondary importance. This said, one viable approximate approach to this problem is to bypass the marginalization entirely, obtain $S_{1:T}$ by solving the inference problem in \eqref{eq:inference} using the Viterbi algorithm, and then promote the $\hat{S}_{1:L}$ mapped from such $S_{1:T}$ as a solution to the decoding problem. 
Projects using this approach have reported good performances \cite{durationviterbi, bioinfomatics}, despite the absence of any theoretical optimality guarantees in terms of \eqref{eq:decoding}. The same approach also extends to other graphic models, such as Bayesian network and conditional random field \cite{Chiron, Bonito}, where the Viterbi is generalized into the max-product algorithm that can find the most-likely path along the graph.

Other approaches strive to account for the state hierarchy to achieve a better approximation by partially marginalizing over the set $S_{1:T}$ that maps to any $\hat{S}_{1:L}$. The surjective mapping from $S_{1:T}$ to $\hat{S}_{1:L}$ necessitates constraints on the inner states structure that explicitly marks the start/end of states at each level.\footnote{Otherwise, the mapping from $S_{1:T}$ to $\hat{S}_{1:L}$ is not proper. An example from nanopore base calling, where the states are the four nucleotides $\{\A,\C,\G,\T \}$, is that $S^1_{1:5}=\{\C,\C,\A,\A,\T \}$ can map to either $\hat{S}_{1:3}=\{\C,\A,\T \}$ or $\hat{S}_{1:4}=\{\C,\A,\A,\T \}$ with $\hat{S}_2=\A$ ending at $t=3$ or $t=4$ alternatively.} In \cite{LTHHMM}, the authors introduce a binary indicator variable $F^r_t$, where $F^r_{t}=1$ indicates the end of a state at a higher level $r-1$ at time $t$, forcing a state transition, and $F_t^r=0$ indicates the continuation of a state.
Let $(S^1_{1:T},F^2_{1:T})$ denote the jointly time-aligned outer states and the second-order indicators. One can then in principle compute the marginal probability $P(S^1_{1:T},F^2_{1:T},X_{1:T})$ from summing $P(S_{1:T},X_{1:T})$ over all $(S_{1:T},F_{1:T})$ that maps to $(S^1_{1:T},F^2_{1:T})$, then $P(\hat{S}_{1:L},X_{1:T})$ by summing $P(S^1_{1:T},F^2_{1:T},X_{1:T})$ over all $(S^1_{1:T},F^2_{1:T})$ that maps to $\hat{S}_{1:L}$. The Marginalized Viterbi algorithm (MVA) was developed in \cite{Hayashi2013} to find the most likely outer state jointly with the indicators, i.e., $\argmax_{S^1_{1:T},F^2_{1:T}}P(S^1_{1:T},F^2_{1:T},X_{1:T})$. The MVA is a re-discovery of the modified Viterbi algorithm of \cite{bioinfomatics}, designed to find the outer states with its `critical edges', conceptually equivalent to the indicators $F^2_{1:T}$.  
The sequence $\hat{S}_{1:L}$ is then simply extracted from the MVA result $(S^1_{1:T},F^2_{1:T})$ and promoted as the decoding solution. It is proven in \cite{bioinfomatics} that such $\hat{S}_{1:L}$ mapped from $(\hat{S}^1_{1:T},F^2_{1:T})$ is the the optimal solution of the decoding problem in \eqref{eq:decoding} only if the mapping is one-to-one (bijective), which is rarely realistic. Proper marginalization over $F^2_{1:T}$ generally remains intractable.

In this paper, we propose an alternative approach for decoding HHMMs that further marginalizes the `critical edges' $F^2_{1:T}$ to approximately compute $\argmax P(\hat{S}_{1:L},X_{1:T})$ in \eqref{eq:inference}, instead of completely bypassing or only partially marginalizing. To achieve this, we extend the pruning-based Beam Search (BS) algorithm to the HHMM. In the following sessions, we first describe the inference methods available for general HHMMs. Then, we outline the fundamental principles behind applying the modified BS algorithms on HHMMs to find the approximated $\hat{S}_{1:L}$. Depending on the pruning approach selected, our proposal yields the Greedy Marginalized BS (GMBS) and the Local Focused BS (LFBS), with the latter having reduced complexity.

\begin{figure}
  \centering
  \includegraphics[width=0.85\columnwidth]{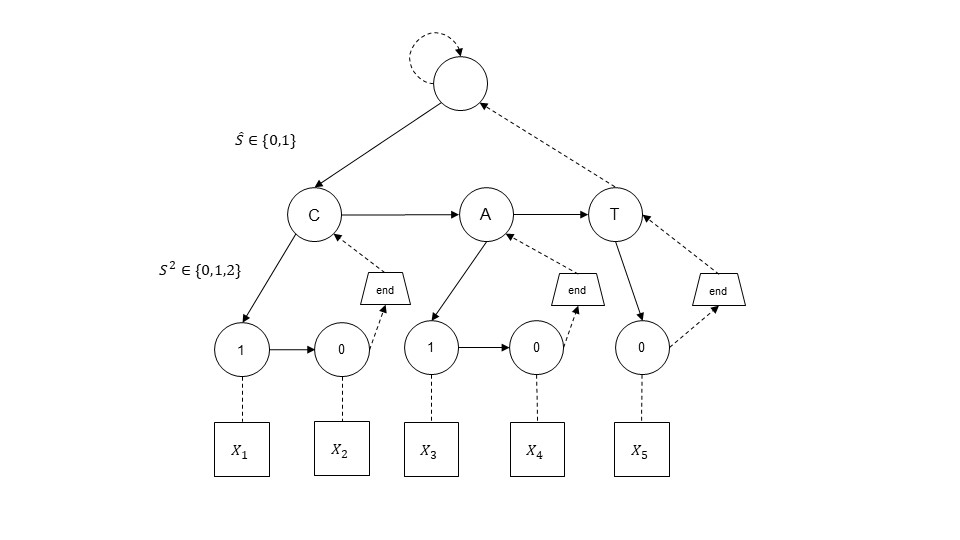}
  \caption{Illustration of a rank $2$ HHMM with $\hat{s}_{1:3}=\{\C,\A,\T \}$ and the inner states $\{1,0\}$, $\{1,0\}$ and $\{0\}$ dominated by each outer state respectively. We define $\mathbb{S}^2_{\text{e}}=\{0\}$ and $\mathbb{S}^2_{\text{c}}=\{1,2\}$ as the end and continuation subset of the inner state space. The dashed arrow indicates when the lower level states end and forces transit in higher level states.}
  \label{HHMM}
\end{figure}
\begin{figure}
  \centering
  \includegraphics[width=0.85\columnwidth]{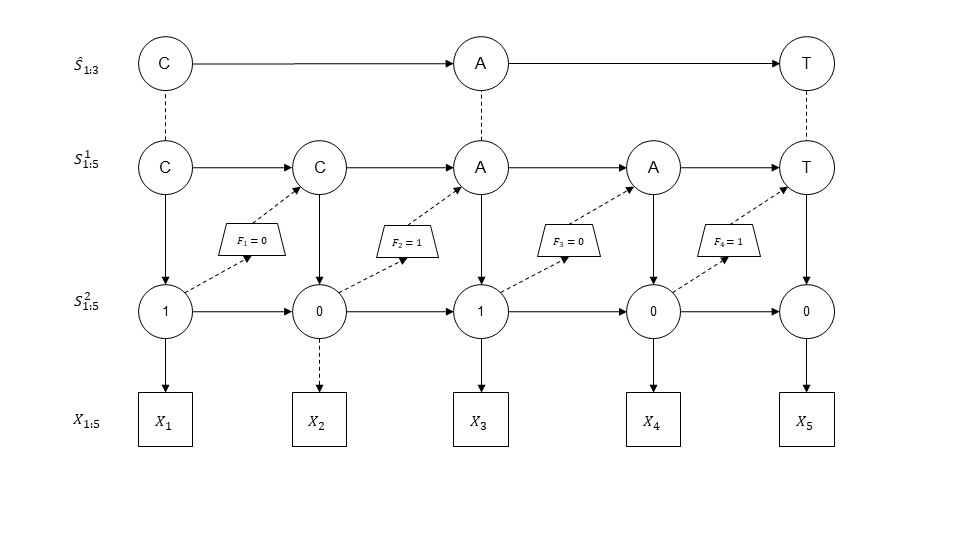}
  \caption{Illustration of a rank $2$ time-alignment of the HHMM shown in Figure \ref{HHMM}. It is reformed as the time-aligned outer states $S^1_{1:5}=\{\C,\C,\A,\A,\T \}$ and inner states $S^2_{1:5}=\{1,0,1,0,0\}$. The indicators $F_2=1$ and $F_4=1$ identified the end of $\hat{S}_1$ at time $2$ and $\hat{S}_2$ at time $4$, which is equivalent to have $S^2_2=0$ and $S_4^2=0$, with end set defined as  $\mathbb{S}^2_{\text{e}}=\{0\}$. Evidently, $F_{t}=1$ whenever $S^2_t\in \mathbb{S}^2_{\text{e}}$.} 
  \label{DBN}
\end{figure}

\section{Hierarchical HMMs}\label{chap2}

For simplicity of notation, we only consider $R=2$ in the remainder of this work, i.e., $r\in\{1,2\}$, because the principles are the same for hierarchies with $R>2$. In order to achieve the surjective mapping from $S_{1:T}$ to $\hat{S}_{1:L}$ similar to the introduction of $F_{1:T}$ in \cite{LTHHMM}, but with a more concise notation, we design the inner state space to consist of two subsets named the \textit{end} and the \textit{continuation} sets, denoted by $\mathbb{S}^2_{\text{e}}$ and $\mathbb{S}^2_{\text{c}}$. If the inner state takes a value from the end subset, the current outer state is forced to end and will transit; otherwise, the outer state continues dwelling.\footnote{One can, for example, relate the end and the continuation subsets to $F^2_{t}$ by considering the end set as the joint state space $\mathbb{S}^2_{\text{e}}=\{\mathbb{S}^2, F=1\}$, and the continuation set as $\mathbb{S}^2_{\text{c}}=\{\mathbb{S}^2, F=0\}$ for some original inner state space $\mathbb{S}^2$.} Figures \ref{HHMM} and \ref{DBN} show an example of a rank $2$ HHMM and its time-aligned representation. For simplicity of exposition, we define $\mathbb{S}\triangleq\{\A,\C,\G,\T \}$ and $\mathbb{S}^2 \triangleq \{1,...,|\mathbb{S}^2|\}$ for all examples in this paper, consistent with our benchmarking experiments using the nanopore base-calling dataset, where $\mathbb{S}$ and $\mathbb{S}^2$ denote the nucleotide bases and an explicit duration variable, respectively, and let $\mathbb{S}^2_{\text{e}} \triangleq \{1\}$ and $\mathbb{S}^2_{\text{c}} \triangleq \{2,...,|\mathbb{S}^2|\}$.

Being an extension of HMMs, the HHMMs also assume the Markov property of all ranked states and the observations. Given $S_{1:t}$ and $X_{1:T}$, one can compute $P(S_{1:t},X_{1:t})$ on the HHMM using the forward algorithm designed for plain HMMs by treating the joint states $S_t$ as a (combined) hidden state. The joint probability 
$P(\hat{S}_{1:l},X_{1:t})$ can, in principle but at considerable complexity, be computed by marginalizing $P(S_{1:t},X_{1:t})$ over the set $\mathbb{B}_t(\hat{S}_{1:l})$ of sequences $S_{1:t}$ that maps to the sequence-aligned outer state sequence $\hat{S}_{1:l}$, according to 
\begin{equation}
P(\hat{S}_{1:l},X_{1:t}) = \sum_{S_{1:t}\in \mathbb{B}_t(\hat{S}_{1:l})}P(S_{1:t},X_{1:t})
\label{psx}
\end{equation}
If we use $\chi_t(\hat{S}_{1:l},i)$ to denote the joint probability of $X_{1:t}$ and $\hat{S}_{1:l}$ with its current inner state $S^2_t=i\in \mathbb{S}^2$ at time $t$ then, naturally, $P(\hat{S}_{1:l},X_{1:t})$ can be rewritten using $\chi_t(\hat{S}_{1:l},i)$ after marginalization over $S^2_t$, i.e.,
\begin{equation}
    P(\hat{S}_{1:l},X_{1:t}) = \sum_{i\in \mathbb{S}^2}\chi_t(\hat{S}_{1:l},i) \, .
    \label{newpsx}
\end{equation}
Combing \eqref{psx} with \eqref{newpsx} yields
\begin{equation*}
\chi_t(\hat{S}_{1:l},i)=\sum_{S_{1:t}\in\mathbb{B}_t(\hat{S}_{1:l}),S^2_t=i}P(S_t=(\hat{S}_l,i),S_{1:t-1},X_{1:t}) \, ,
\label{setB}
\end{equation*}
where $\chi_t(\hat{S}_{1:l},i)$ is computed as the sum over a subset of $\mathbb{B}_t(\hat{S}_{1:l})$ where the current inner state $S^2_t=i$. It can be shown (see Appendix A for detailed proofs) that the probability $\chi_t(\hat{S}_{1:l},i)$ can be computed recursively from $\chi_{t-1}$ as follows:
\begin{align}
\begin{split}
\chi_t(\hat{S}_{1:l},i)&=\sum_{j \in \mathbb{S}^2_{\text{e}}}P\Big(S_t=(\hat{S}_l,i),X_t|S_{t-1}=(\hat{S}_{l-1},j)\Big)\chi_{t-1}(\hat{S}_{1:l-1},j) \\
& \qquad + \sum_{k\in\mathbb{S}^2_{\text{c}}}P\Big(S_t=(\hat{S}_l,i),X_t|S_{t-1}=(\hat{S}_l,k)\Big)\chi_{t-1}(\hat{S}_{1:l},k) \, ,
\end{split}
\label{twosum}
\end{align}
in which the two terms represent two sources of the sequence $\hat{S}_{1:l}$ with a specific inner state $i$ emerging at time $t$: the first term indicates that the sequence $\hat{S}_{1:l-1}$ existed at $t-1$ and extended a new outer state $S^1_t=\hat{S}_{l}$  at time $t$, while the second term indicates the same sequence $\hat{S}_{1:l}$ that existed at $t-1$ and continued dwelling at time $t$. Marginalizing over the set $\mathbb{B}_t(\hat{S}_{1:l})$ entails recursively marginalizing over the two terms related to $\hat{S}_{1:l-1}$ and $\hat{S}_{1:l}$. 

We can further simplify the model with the Markov assumption
\begin{equation*}
P(S_t,X_t|S_{t-1})=P(X_t|S_t,S_{t-1})P(S_t|S_{t-1})=P(X_t|S_t)P(S_t|S_{t-1}),
\end{equation*}
where $P(S_t|S_{t-1})$ is the HHMM transition probabilities, and $P(X_t|S_t)$ is the emission probability of observation $X_t$ given the joint state $S_t=(S^1_t,S^2_t)$.\footnote{We use the first-order Markov assumption here for notational simplicity only. The argument is easily extended to higher-order Markov processes, and in the benchmark experiments described later, we assumed a fifth-order Markov chain, e.g., $P(S_t,X_t|S^2_{t-1},\hat{S}_{l-4:l})=P(X_t|\hat{S}_{l-4:l})P(S_t|\hat{S}_{l-4:l})$ or $P(S_t,X_t|S^2_{t-1},\hat{S}_{l-5:l-1})=P(X_t|\hat{S}_{l-5:l-1})P(S_t|\hat{S}_{l-5:l-1})$.} On a hybrid construction of HHMMs and Neural Networks (NNs), the probability $P(X_t|S_t)$ may be replaced by a score $\phi(S_t|X_{t1:t2})$, which is the output of an NN assigned to the joint state $S_t$ given a segment of observations as inputs\cite{Kamaric1999,Lokatt,CNN-HMM,DNN-HMM}.
Finally, we achieve a recursive update of $\chi_t(\hat{S}_{1:l},i)$ as
\begin{align}
\begin{split}
\chi_t(\hat{S}_{1:l},i) & =  P\Big(X_t|S_t=(\hat{S}_l,i)\Big)\Bigg(\sum_{j\in\mathbb{S}^2_{\text{e}}}P\Big(S_t=(\hat{S}_l,i)|S_{t-1}=(\hat{S}_{l-1},j)\Big)\chi_{t-1}(\hat{S}_{1:l-1},j)\\
& \qquad \qquad \qquad \qquad \qquad \quad +\sum_{k\in\mathbb{S}^2_{\text{c}}}P\Big(S_t=(\hat{S}_l,i)|S_{t-1}=(\hat{S}_l,k)\Big)\chi_{t-1}(\hat{S}_{1:l},k)\Bigg)\, .
\end{split}
\label{chi}
\end{align}

The likelihood of any given $(\hat{S}_{1:L},S^2_T)$ and $X_{1:T}$ can, in principle, be computed by applying the recursion in \eqref{chi} for $t=1,...,T$. Together with \eqref{newpsx}, we can reformulate the decoding objective with $\chi_T$ as follows:
\begin{equation*}
    \hat{S}^*_{1:L^*} = \argmax_{\hat{S}_{1:L},L\leq T}P(\hat{S}_{1:L},X_{1:T}) =\argmax_{\hat{S}_{1:L},L\leq T}\sum_{i\in \mathbb{S}^2}\chi_T(\hat{S}_{1:L},i) .
\end{equation*}

Ideally, one would obtain the optimal solution $\hat{S}^*_{1:L^*}$ by computing and sorting the likelihood $P(\hat{S}_{1:L},X_{1:T})$ for all candidate sequences $\hat{S}_{1:L}$ and all $L\leq T$. However, this is computationally infeasible as the total number of candidates, i.e., $\sum_{1\leq L\leq T}|\mathbb{S}|^L$, scales exponentially in $T$. Nevertheless, we can approximate $\hat{S}^*_{1:L^*}$ with tractable algorithms such as the BS algorithm, which only evaluates the recursion in \eqref{chi} over a pruned candidate group. In the next chapter, we modify and generalize the standard BS algorithm to HHMMs.

\section{Marginalized Beam Search Algorithms }

With the term `beam search' first formalized in \cite{BS}, the BS algorithm offers reduced computational complexity for decoding HMMs \cite{improveBS,viterbiBS}  and conditional random fields \cite{BScrf,BSnayesiannetwork} compared to other well-known algorithms, such as the Viterbi algorithm. To achieve this, the BS optimizes a best-first search approach by maintaining a fixed beam-width $W$ and tracking the $W$ most likely state sequences, called \textit{beams}, throughout the recursive forward search steps. At the start of each time step, the BS expands all possible next steps of each existing beam to create a set of \textit{leaf beams}. Each leaf beam is assigned a score based on its parent beam and the current observation. The BS then sorts all leaf beams according to their scores and keeps the top-$W$ as the new set of beams for the next step while pruning the rest. By recursively performing the expanding and pruning at each time step, the standard BS approximates the optimal solution of inference or decoding problems in a greedy manner. 

However, in the context of decoding HHMMs, the BS requires an additional \textit{marginalization} step before the pruning to maintain the recursive marginalization in \eqref{chi}. We refer to the BS algorithm with this additional step as the marginalized BS and introduce two variations based on different pruning steps in the following sections, the Greedy Marginalized BS and the Local Focused BS.

\subsection{Greedy Marginalized Beam Search}\label{sectionGMBS}
The GMBS algorithm accommodates $W$ beams, each capturing a \emph{sequence-aligned} outer state sequence. We define $\mathbb{B}^{\text{BS}}_{t-1}\triangleq\{\hat{S}_{1:l^1_{t-1}},...,\hat{S}_{1:l^W_{t-1}}\}$ to be the collection of candidate sequences captured at time $t-1$ by the $W$ beams and denote the beams by $(\hat{S}_{1:l^w_{t-1}},X_{1:t-1})$ for $w=1,...,W$. At the beginning of time step $t$, the beams are expanded into a group of leaf beams that represent all possible next steps of $\mathbb{B}^{\text{BS}}_{t-1}$ in a union ${\mathbb{B}^{\text{BS}}_{t-1} \cup \bar{\mathbb{B}}_{t}}$. Specifically, the \textit{extension} set $\bar{\mathbb{B}}_t$ is the collection of the sequences $\hat{S}_{1:(l^w_{t-1}+1)}$ for $w=1,...,W$ created by extending a new outer state $\hat{S}_{(l^w_{t-1}+1)}\in \mathbb{S}$ at the end of $\hat{S}_{1:l^w_{t-1}}$, captured in the leaf beams $(\hat{S}_{1:(l^w_{t-1}+1)},X_{1:t})$. Meanwhile, there are leaf beams $(\hat{S}_{1:l^w_{t-1}},X_{1:t})$ that capture the same sequences $\hat{S}_{1:l^w_{t-1}}\in \mathbb{B}^{\text{BS}}_{t-1}$ as represented in the original $W$ beams. We will refer to $\mathbb{B}^{\text{BS}}_{t-1}$ as the  \textit{continuation} set in contrast to the extension set.

Essentially, the extension and the continuation sets correspond to the two sources (i.e., sums) in \eqref{twosum} and \eqref{chi}. Therefore, given the scores of all existing beams at $t-1$, one can update the score of a specific leaf beam $(\hat{S}_{1:l},X_{1:t})$ with any sequence $\hat{S}_{1:l}\in {\mathbb{B}^{\text{BS}}_{t-1} \cup \bar{\mathbb{B}}_{t}}$ analogously to \eqref{chi} as
\begin{align}
\begin{split}
\bar{\chi}_t(\hat{S}_{1:l}) & = \sum_{i\in\mathbb{S}^2}\bar{\chi}_t(\hat{S}_{1:l},i)  \\
& = \sum_{i\in\mathbb{S}^2}P\big(X_t|S_t=(\hat{S}_{l},i)\big)\Big( \sum_{j\in\mathbb{S}^2_{\text{e}}}P\big(S_t=(\hat{S}_{l},i)|S_{t-1}=(\hat{S}_{l-1},j)\big)\bar{\chi}_{t-1}(\hat{S}_{1:l-1},j)\\
& \qquad \qquad \qquad \qquad \qquad \qquad + \sum_{k\in\mathbb{S}^2_{\text{c}}}P\big(S_t=(\hat{S}_{l},i)|S_{t-1}=(\hat{S}_{l},k)\big)\bar{\chi}_{t-1}(\hat{S}_{1:l},k)\Big) ,
\end{split}
\label{GMBSeq}
\end{align}
where we use the score $\bar{\chi}_t(\hat{S}_{1:l})$ and $\bar{\chi}_t(\hat{S}_{1:l},i)$ to replace the probabilities $P(\hat{S}_{1:l},X_{1:t})$ and $\chi_t(\hat{S}_{1:l},i)$ respectively because they are not properly normalized due to the subsequent pruning step. As the GMBS algorithm operates on a pruned state space, it is possible that certain leaf beams may have both sources contained within the beams, while others may lack one of the sources, i.e., $\bar{\chi}_{t-1}(\hat{S}_{1:l-1},j)=0$ if $\hat{S}_{1:l-1}\notin\mathbb{B}^{\text{BS}}_{t-1}$ and $\bar{\chi}_{t-1}(\hat{S}_{1:l},k)=0$ if $\hat{S}_{1:l}\notin\mathbb{B}^{\text{BS}}_{t-1}$. 

In the practical implementation, the computation of \eqref{GMBSeq} is decomposed in two consecutive steps: expansion and marginalization. The expansion step is similar to the standard BS algorithm and involves computing scores of unmarginalized leaf beams, here referred to as the vanilla leaf beams, that capture the extension set $\bar{\mathbb{B}}_t$, and the continuation set $\mathbb{B}^{\text{BS}}_{t-1}$ separately, with $|\bar{\mathbb{B}}_t|=W|\mathbb{S}|$ and $|\mathbb{B}^{\text{BS}}_{t-1}|=W$. The score of the vanilla leaf beam representing $\hat{S}_{1:(l^w_{t-1}+1)}\in\bar{\mathbb{B}}_t$ is calculated by the first term in \eqref{GMBSeq}, i.e.,
\begin{equation*}
     \tilde{\chi}^\text{e}_t(\hat{S}_{1:(l^w_{t-1}+1)}) = \sum_{i\in\mathbb{S}^2}P\big(X_t|S_t=(\hat{S}_{(l^w_{t-1}+1)},i)\big)\sum_{j\in\mathbb{S}^2_{\text{e}}}P\big(S_t|S_{t-1}=(\hat{S}_{l^w_{t-1}},j)\big)\bar{\chi}_{t-1}(\hat{S}_{1:l^w_{t-1}},j)\; \, ,
\end{equation*}
while the score of the vanilla leaf beam representing $\hat{S}_{1:l^w_{t-1}}\in\mathbb{B}^{\text{BS}}_{t-1}$ is calculated by the second term in \eqref{GMBSeq}, i.e.,
\begin{equation*}
     \tilde{\chi}^\text{c}_t(\hat{S}_{1:l^w_{t-1}}) = \sum_{i\in\mathbb{S}^2}P\big(X_t|S_t=(\hat{S}_{l^w_{t-1}},i)\big)\sum_{k\in\mathbb{S}^2_{\text{c}}}P\big(S_t|S_{t-1}=(\hat{S}_{l^w_{t-1}},k)\big)\bar{\chi}_{t-1}(\hat{S}_{1:l^w_{t-1}},k) \; \, .
\end{equation*}
The marginalization step then merges the vanilla leaf beams into a group of leaf beams representing the union ${\mathbb{B}^{\text{BS}}_{t-1} \cup \bar{\mathbb{B}}_{t}}$. It is important to note that the union size may be smaller, i.e., $|{\mathbb{B}^{\text{BS}}_{t-1} \cup \bar{\mathbb{B}}_{t}}|\leq  W\times(1+|\mathbb{S}|)$, due to the possible existence of redundant pairs of vanilla leaf beams representing two sources for the same leaf beam. The GMBS, therefore, needs to merge the redundant vanilla leaf beams into one leaf beam, which captures $\hat{S}_{1:l}\in {\mathbb{B}^{\text{BS}}_{t-1} \cap \bar{\mathbb{B}}_{t}}$ with a score that is the sum of the scored of the redundant vanilla leaf beams, i.e.,
\begin{equation*}
    \bar{\chi}_t(\hat{S}_{1:l})=\tilde{\chi}^{\text{e}}_t(\hat{S}_{1:l})+\tilde{\chi}^{\text{c}}_t(\hat{S}_{1:l})\; .
\end{equation*}
The non-redundant vanilla leaf beams are promoted directly to the leaf beams, inheriting the same score, i.e., $\bar{\chi}_t(\hat{S}_{1:l})=\tilde{\chi}^\text{e}_t(\hat{S}_{1:l})$ for $\hat{S}_{1:l}\in\bar{\mathbb{B}}_{t} \setminus \mathbb{B}^{\text{BS}}_{t-1} $, and $\bar{\chi}_t(\hat{S}_{1:l})=\tilde{\chi}^\text{c}_t(\hat{S}_{1:l})$ for $\hat{S}_{1:l}\in {\mathbb{B}^{\text{BS}}_{t-1}\setminus\bar{\mathbb{B}}_{t}}$. This merging step implicitly marginalizes over all possible ways of aligning the state sequences to the observations, albeit only over the set of possibilities represented by the kept beams.

However, finding the redundant vanilla leaf beams can pose a significant challenge in terms of both memory and computational resources, especially as the length $l$ increases. Keeping all sequences in $\bar{\mathbb{B}}_t$ and $\mathbb{B}^\text{BS}_t$ in memory and then comparing them individually is not a practical approach. In order to mitigate this issue, we propose an identifier (ID) based system, which will be discussed in the upcoming section. This system also ensures efficient marginalization.

After the expansion and the marginalization steps, the GMBS proceeds to perform the pruning step, which keeps only the top-$W$ highest-scored leaf beams from among the leaf beams representing $\hat{S}_{1:l}\in \{\mathbb{B}^{\text{BS}}_{t-1} \cup \bar{\mathbb{B}}_{t}\}$:
\begin{equation*}
    \mathbb{B}^{\text{BS}}_t = \argmax_{\underbrace{\hat{S}_{1:l_t^1},...,\hat{S}_{1:l_t^W}}_{\text{keep top-}W}}\bar{\chi}_t(\hat{S}_{1:l}) \quad \forall \; \hat{S}_{1:l}\in \{\mathbb{B}^{\text{BS}}_{t-1} \cup \bar{\mathbb{B}}_{t}\} \; .
\end{equation*}
The $W$ surviving leaf beams are then promoted to the new set of beams representing $\mathbb{B}^{\text{BS}}_t$ at time $t$, initiating the next step of the recursive loop. 

\subsubsection{Marginalization and Back-tracking with the ID System}\label{IDs}
The detection of redundant vanilla leaf beams is achieved by leveraging the commonality of their last outer state $\hat{S}_l$, and tracking back to the same ancestor beam that represents $\hat{S}_{1:l-1}$ at an earlier time. Despite storing and comparing the last state $\hat{S}_l$ being a straightforward task, tracking the ancestor beams back in time can be challenging. We therefore propose an ID system that serves as a guidance mechanism to identify the common ancestor beam of redundant leaf beams in a computationally efficient manner. 

During the expansion step at time $t$, each vanilla leaf beam is assigned an ID and a parent ID. More specifically, the vanilla leaf beam $(\hat{S}_{1:l},X_{1:t})$ capturing the extension set $\hat{S}_{1:l}\in \bar{\mathbb{B}}_t$ is assigned a newly generated ID, which incorporates the current timestamp $t$, and a parent ID that is the ID of the beam $(\hat{S}_{1:l-1},X_{1:t-1})$ from which it was expanded. Meanwhile, a vanilla leaf beam $(\hat{S}_{1:l},X_{1:t})$ capturing the continuation set $\hat{S}_{1:l}\in \mathbb{B}^{\text{BS}}_{t-1}$ retains both the ID and the parent ID of the original beam $(\hat{S}_{1:l},X_{1:t-1})$ without IDs being generated. 

\begin{figure}
  \centering
  \includegraphics[width=0.85\columnwidth]{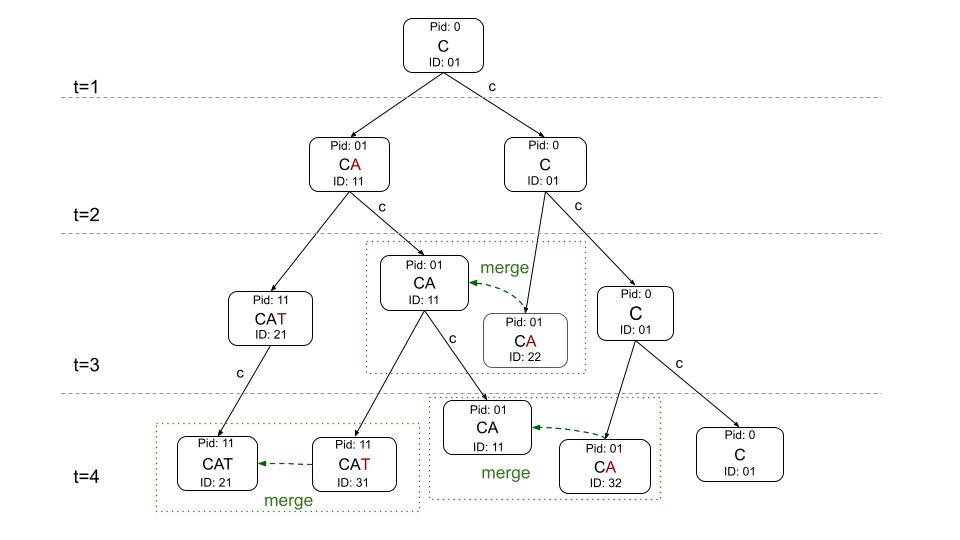}
  \caption{In this depiction, each vanilla leaf beam is represented as a rectangular shape, displaying its unique ID at the bottom and its parent ID (Pid) at the top. We use a two-digit ID, where the left digit is the timestamp and the right digit is unique for each vanilla leaf beam. The inner states have been omitted, and the Pid at $t=0$ is fixed as $0$. A small `c'  is put next to the arrow to indicate the continuation set. The temporal progression of the algorithm is exhibited for $t=1,2,3$, shown in top-down order. At $t=2$, two redundant vanilla leaf beams with ID $11$ and ID $22$ share the same Pid and last outer state; hence they are merged, preserving the ID $11$ and the Pid $01$. At $t=3$, two pairs of redundant vanilla leaf beams are merged: those with ID $21$ and ID $31$ unite, preserving the ID $21$ and Pid $11$, and those with ID $11$ and ID $32$ merge, preserving the ID $11$ and Pid $01$. By $t=3$, the GMBS generates three beams representing all possible sequence-aligned outer states of various lengths: $\{\C\}$, $\{\C,\A\}$, and $\{\C,\A,\T\}$. Note that the merging logic can be implemented only using the Pid and the last state, obviating the need for full sequence matching.}
  \label{MBS}
\end{figure}

During the subsequent marginalization step, the algorithm first group all vanilla leaf beams by their parent IDs, which can be achieved efficiently by sorting the parent IDs. Then for the vanilla leaf beams sharing the same parent ID, the algorithm compares their last outer state to detect redundancies. Upon detection, the algorithm merges the redundant vanilla leaf beams into a new leaf beam with a score given by the sum of the vanilla leaf beams. It assigns the newly created leaf an ID according to the following rules: The resulting leaf beam retains both the ID and parent ID of the vanilla leaf beam capturing the continuation set, i.e., $\hat{S}_{1:l}\in \mathbb{B}^{\text{BS}}_{t-1}$, whose ID has the earlier timestamp. We illustrate the ID system in Figure \ref{MBS} with an example for inferring a sequence $\{\C,\A,\T\}$ using the expansion and marginalization steps in the GMBS, leaving out the pruning step.

\begin{figure}
  \centering
  \includegraphics[width=0.85\columnwidth]{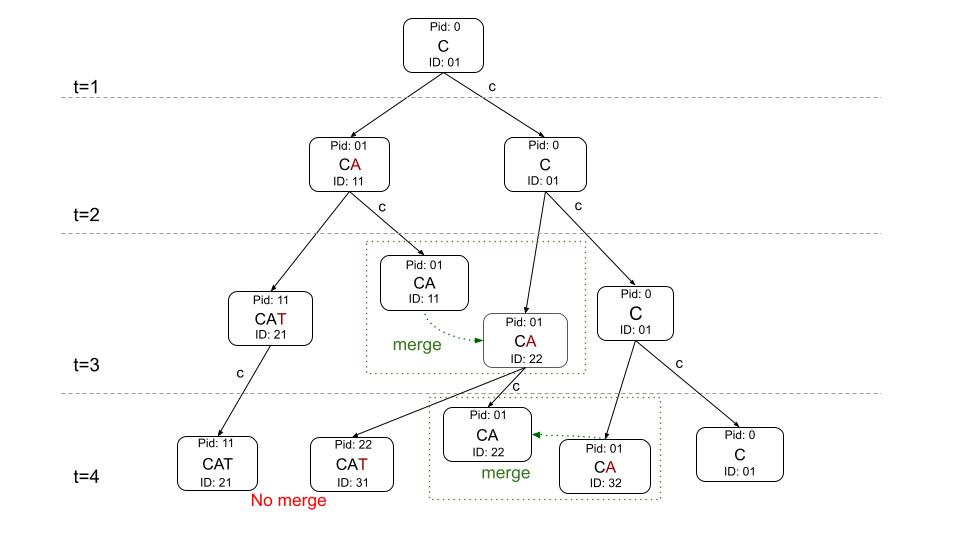}
  \caption{Illustration of missed detections of possible mergers when preserving the ID with the later timestamp in the same example shown in Figure \ref{MBS}. At $t=2$, the leaf beam instead preserves ID $22$. Consequently, at $t=3$, the vanilla leaf beam with ID $31$ has Pid $22$, which is not the same Pid shared by its redundant vanilla leaf beam with ID $21$. Therefore, the redundant pair, i.e., ID $21$ and ID $31$, can not be properly marginalized at time $3$.}
  \label{MBSwrong}
\end{figure}

Crucially, the resulting leaf beam must carry the earlier timestamp ID to ensure the detection of redundant vanilla leaf beams in the future. Conversely, a redundant vanilla leaf beam capturing the extension set with the later timestamp (the current $t$) will not be the ancestor beam for any vanilla leaf beam at time $t+1$; hence it would not be merged in any future step. Figure \ref{MBSwrong} demonstrates a case where redundancies are missed when the new leaf beam carries the later timestamp ID. 

In the pruning step, the GMBS keeps the top-$W$ leaf beams while maintaining their IDs, parent IDs and the last outer states. At time $t=T$, the GMBS performs a back-tracking procedure to reconstruct $\hat{S}_{1:L}$ with the help of the ID system. Starting with the most likely beam at time $T$, the GMBS traces back to its ancestor beam at a specific earlier time suggested by its parent ID, reversely concatenating their stored outer states. Then this procedure repeats, tracking the ancestor beams suggested by the parent IDs of the retrieved ancestor beams and concatenating their outer states until the parent ID timestamp is $0$. Eventually, the sequence $\hat{S}_{1:L}$ is obtained by concatenating all stored outer states in the ancestor beams.\footnote{Despite the timestamps carrying the meaning of a `critical edge' when it is created, they are not the `critical edges' for the retrieved $\hat{S}_{1:L}$ due to the marginalization step.}

The formalized approach of the GMBS is shown in Algorithm \ref{GMBS}. With an infinite width $W$, the GMBS would compute the optimal decoding solution. Therefore, it is intuitive to select a large value of $W$. However, the GMBS entails sorting of up to $W\times(|\mathbb{S}|+1)$ items, which has a time complexity of $\mathcal{O}\big(W|\mathbb{S}|\log(W|\mathbb{S}|)\big)$ in a sequential implementation and of $\mathcal{O}\big(\log^2 \big(W|\mathbb{S}|\big)\big)$ in a parallel implementation assuming a parallel sorting algorithm such as a bitonic sorter \cite{bitonicsorter}. The expansion step requires computing the probabilities over the inner states, i.e., computing $\chi_t$ in \eqref{twosum} for all $i\in \mathbb{S}^2$ and all beams, which has a time complexity of $\mathcal{O}(W|\mathbb{S}^2|^2)$ in a sequential implementation and of $\mathcal{O}(|\mathbb{S}^2|^2)$ in a parallel implementation. Therefore, the total time complexity of the sequential implementation is $\mathcal{O}\big(TW|\mathbb{S}^2|^2+TW|\mathbb{S}|\log(W|\mathbb{S}|)\big)$, and the time complexity of a parallel implementation with at least $W$ threads is $\mathcal{O}\big(T|\mathbb{S}^2|^2+T\log^2(W|\mathbb{S}|)\big)$. To eliminate this costly sorting operation, we propose the LFBS algorithm that requires sorting only $W$ lists, each of $|\mathbb{S}|+1$ items.

\subsection{Local Focus Beam Search}
The LFBS provides a computationally efficient alternative to the GMBS, and involves the same expansion and marginalization step but a modified pruning step. With a fixed \textit{focus-length} $K$, the LFBS algorithm maintains $M=|\mathbb{S}|^K$ beams at each step $t$. Each beam captures the most probable sequence that ends with a particular sequence of $K$ consecutive outer states. In other words, the LFBS only focuses locally on the last $K$ outer states of each candidate $\hat{S}_{1:l}$ via the implementation of the pruning step. 

We denoted the $m$th beam at time $t-1$ as $(\hat{S}_{1:l^m_{t-1}},X_{1:t-1})$ for $m\in\{1,...,M\}$, such that $\hat{S}_{(l^m_{t-1}-K+1):l^m_{t-1}}=A_m \in \mathbb{S}^{K}$. 
At the start of step $t$, the existing beams expand into leaf beams with scores computed by \eqref{GMBSeq} using the same expansion and marginalization steps in the GMBS. 
When a vanilla leaf beam $(\hat{S}_{1:(l^m_{t-1}+1)},X_{1:t})$ is created at time $t$ by extending the beam $(\hat{S}_{1:l^m_{t-1}},X_{1:{t-1}})$ that existed at time $t-1$, the LFBS ignores $\hat{S}_{l^m_{t-1}-K+1}$ and shift the focus on the last $K$ outer states of the new leaf beam, i.e., $\hat{S}_{(l^m_{t-1}-K+2):(l^m_{t-1}+1)}=A_n$ for certain $n\in\{1,...,M\}$.\footnote{For example, both beams that end with $\{\G,\C,\A\}$ and $\{\A,\C,\A\}$ can expand to leaf beams that end with $\{\C,\A,\T\}$ by dropping the first $\G$ or the first $\A$, respectively, and extending with the $\T$ in the end.} Here $n=m$ if the last $K+1$ outer states $\hat{S}_{(l^m_{t-1}-K+1):(l^m_{t-1}+1)}$ are all identical. For any specific $n$ and the sequence of $K$ outer states $A_n$, there are $|\mathbb{S}|$ vanilla leaf beams ending with $A_n$, since there are $|\mathbb{S}|$ values of $\hat{S}_{l^m_{t-1}-K+1}$. Including the one vanilla leaf beam created from continuing the original beam $(\hat{S}_{1:l^n_{t-1}},X_{1:{t-1}})$, there are a total number of $|\mathbb{S}|+1$ vanilla leaf beams capturing sequences ending with the same $A_n$. Moreover, as the beams encode their last $K$ outer states with their index $m$, the indexes of the $|\mathbb{S}|+1$ source beams for each $n$ can be predetermined and remain the same throughout the algorithm. 
While this does not reduce the time complexity compared to a GMBS implementation with $W = M$ beams, it facilitates static memory allocations and access patterns for the list of beams and potentially large practical benefits in terms of run-time in an optimized software.

\begin{figure}
  \centering
  \includegraphics[width=0.85\columnwidth]{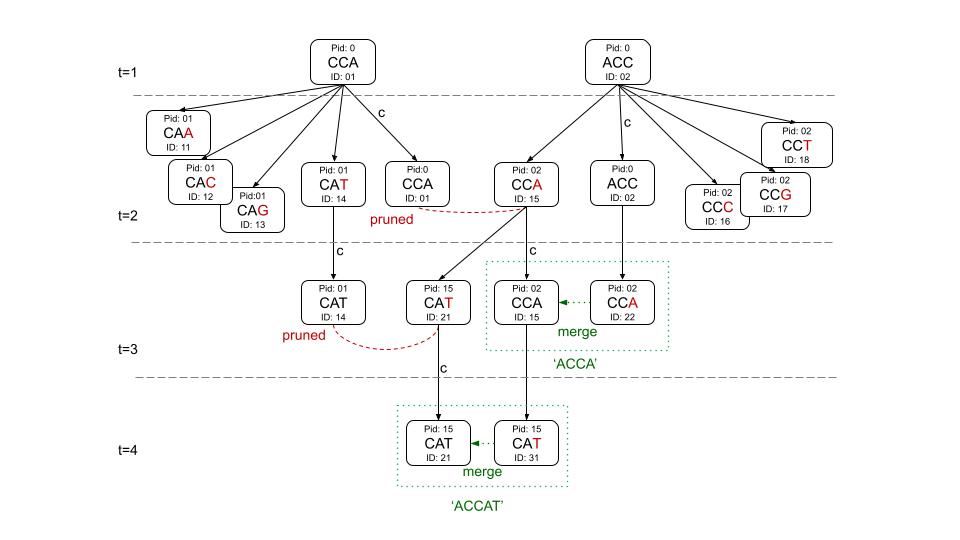}
  \caption{An implementation of the LFBS with $K=3$, given the target sequence $\{\A,\C,\C,\A,\T\}$. The inner states transitions and leaf beams with low scores are omitted in this illustration for simplicity. Only at $t=1$ all vanilla leaf beams are listed: $5$ vanilla beams from each original beam. At $t=1$ the leaf beams with ID $01$ (representing $\{\C,\C,\A\}$) and ID $15$ (representing $\{\A,\C,\C,\A\}$) both represent the last $3$ outer states $\{\C,\C,\A\}$. The LFBS prunes the leaf beam with ID $01$ (assuming a lower score) and keeps the leaf beam with ID $15$. At $t=2$, the redundant leaf beams ID $15$ and ID $22$ (representing $\{\A,\C,\C,\A\}$) are detected and merged. In the following pruning step, the leaf beam with ID $21$ is kept while the leaf beam with ID $14$ is pruned. At $t=3$, leaf beams with ID $21$ and ID $31$ are merged (representing $\{\A,\C,\C,\A,\T\}$). In the end, the LFBS has computed $P(\{\A,\C,\C,\A,\T\},X_{1:3})$.}
  \label{LFBSDNA}
\end{figure}

The expansion step results in $M$ groups of vanilla leaf beams, such that the $m$th group captures $|\mathbb{S}|+1$ sequences ending with the same $K$ outer states $A_m$. Therefore, the LFBS only needs to verify redundant vanilla leaf beams within each group during the marginalization step and, similarly, perform the pruning step within each group afterwards. Specifically, during the pruning step, the LFBS algorithm sorts $M$ lists of scores of all groups, each of which contains at most $|\mathbb{S}|+1$ leaf beams, and then promotes the highest-scored one of the $m$th group to the $m$th beam in the time $t$ while pruning the rest of the group, i.e.,
\begin{equation*}
    (\hat{S}_{1:l^m_t},X_{1:t}) = \argmax_{(\hat{S}_{1:l},X_{1:{t}})} \bar{\chi}_t(\hat{S}_{1:l}) \quad \forall \hat{S}_{(l-K+1):l}=A_m\; .
\end{equation*}
This modified pruning step of the LFBS returns $M$ beams, each representing the most likely $\hat{S}_{1:l^m_t}$ ending with a specific $A_m$ at time $t$. Figure \ref{LFBSDNA} illustrates an implementation of the LFBS algorithm with the target sequence $\{\A,\C,\C,\A,\T\}$. 

In the end, by employing the same back-tracking strategy as in the GMBS, the LFBS reconstructs $\hat{S}_{1:L}$ as an approximation of the optimal decoding solution. The LFBS algorithm is formalized in Algorithm \ref{LFBS}. 
Since the LFBS performs the marginalization and pruning within groups of size at most $|\mathbb{S}|+1$, it does not face the problem of a costly sorting algorithm as in the GMBS. Combined with the expansion step, the total time complexity of the LFBS in a sequential implementation is $\mathcal{O}\big(TM|\mathbb{S}^2|^2+T|\mathbb{S}|\log(|\mathbb{S}|)\big)$, while in a parallel implementation is $\mathcal{O}\big(T|\mathbb{S}^2|^2+T\log^2(|\mathbb{S}|)\big)$. However, it requires $M = |\mathbb{S}|^K$ threads for parallel computing, each thread for a beam, which places a high demand on parallelism. 

We end by pointing out that it follows directly by the analysis of Section~\ref{chap2} that both the GMBS and the LFBS become optimal solutions to the decoding as $W$ and $K$, respectively, grow (very) large. The difference between the two marginalized BS algorithms lies in their respective strategies to select a small but representative subset of candidate beams at each recursion step to reach a given complexity budget. While the beams of the LFBS may not be as representative of the terms in \eqref{eq:decoding} that contribute most to the overall score, the improved parallelism will practically allow for more beams at the same complexity. To assess the relative merits of the respective strategies, we turn next to empirical benchmarking experiments.

%% file: results.tex
\section{Benchmarking Experiments and Results}\label{results}
\subsection{Benchmarking Experiments with the EDHMM}
We conducted two experiments to assess the effectiveness and efficiency of the GMBS and LFBS algorithms in different scenarios. In the first experiment, we simulated the observation and the state sequences using an arbitrary explicit duration hidden Markov model (EDHMM) \cite{EDHMM,durationmodelHMM}, also referred to as Hidden semi-Markov models~\cite{semiHMM}. For the second experiment, we utilized the nanopore base calling dataset from our previous work on the Lokatt model, a hybrid model composed of an EDHMM and a neural network (NN) \cite{Lokatt}. Both experiments involve the EDHMM, which contains an inner state known as the duration variable, denoted by $d$. This variable indicates the remaining time steps that an outer state will be dwelling on the time-aligned sequence, including the present time. The inner state space is therefore defined as $\{1,2,...,D\}$, where $D$ is the maximum (explicit) dwelling time, and can be divided into two subsets: the end subset $\mathbb{S}^2_{\text{e}}=\{1\}$ and the continuation subset $\mathbb{S}^2_{\text{c}}=\{2,3,..., D\}$. By definition, $d$ can only transit to $d-1$ for $1< d < D$ modelling a countdown to an outer transition, while $d=1\in\mathbb{S}^2_{\text{e}}$ forces an end to the outer state. In our project, we also allowed the self-transition at $d=D$ such that the maximum dwell-time of the outer state can approach infinity with a geometrically decaying probability \cite{Lokatt}. 

On the EDHMM, benchmarking against the Viterbi algorithm is equivalent to benchmarking against the MVA or the modified Viterbi algorithms. With the duration state, the mapping from the all-state sequences $S_{1:T}$  to $(S^1_{1:T},F^2_{1:T})$ is one-to-one (bijective). Thus computing $P(S^1_{1:T},F^2_{1:T})$, which required marginalizing over all $(S_{1:T},F_{1:T})$ that maps to $(S^1_{1:T},F^2_{1:T})$, now equals computing $P(S_{1:T},X_{1:T})$. Consequently, the decoding solution provided by the MVA \cite{Hayashi2013}, or the modified Viterbi algorithm in \cite{bioinfomatics}, is identical to the result obtained by the Viterbi algorithm applied directly to joint states of the HHMM \cite{LTHHMM}.

We benchmarked the GMBS and LFBS against the Viterbi algorithm using various configurations in both experiments. Specifically, we implemented the GMBS with beam-width values of 8, 64, 256, and 512, while the LFBS was tested with focus-length values of 5, 6, and 7. All methods were implemented in a highly parallelized form on an NVIDIA V$100$ GPU card.

To assess the decoding performance, we measured the Levenshtein distance \cite{stringmatch} between the decoding solution to a ground truth reference sequence. The Levenshtein distance is the minimal number of single-state edits (insertions, deletions or substitutions) required to change the former to the latter. The sequence accuracy, or the identity score in nanopore base calling, is the ratio of the number of matched outer states to the total number of alignment areas and is calculated as follows:
\begin{equation*}
    \text{accuracy} = \frac{\text{matches}}{\text{matches}+\text{mismatches}+\text{insertion}+\text{deletions}}\; .
\end{equation*}
We define the alignment area to begin at the first matched state and to end at the last matched state, to handle the base calling dataset where the reference is the entire genome (approximately $5$ millions nucleotides) and is much longer than the reads \cite{Lokatt}.

In addition, we also reported decoding performances on areas consisting of $k$ consecutive identical outer states, referred to as the `$k$-mer homopolymers (HPs)' borrowed from the bioinfomatics literature. Decoding these HPs has long been recognized as challenging for many decoding methods, especially in base calling projects where the HPs have random dwelling times and share nearly identical measurements among their component bases. We assessed HP decoding performance in terms of the number of correctly decoded HPs and the number of HPs covered in the mapped reference area, as well as their ratio as the accuracy. All codes and data used in the experiments can be found at the MBS git repository: \hyperlink{https://github.com/chunxxc/MBS.git}{https://github.com/chunxxc/MBS.git}.

\subsection{Simulation Experiment and Results}\label{exp1}
\begin{figure}
  \centering
  \includegraphics[width=0.85\columnwidth]{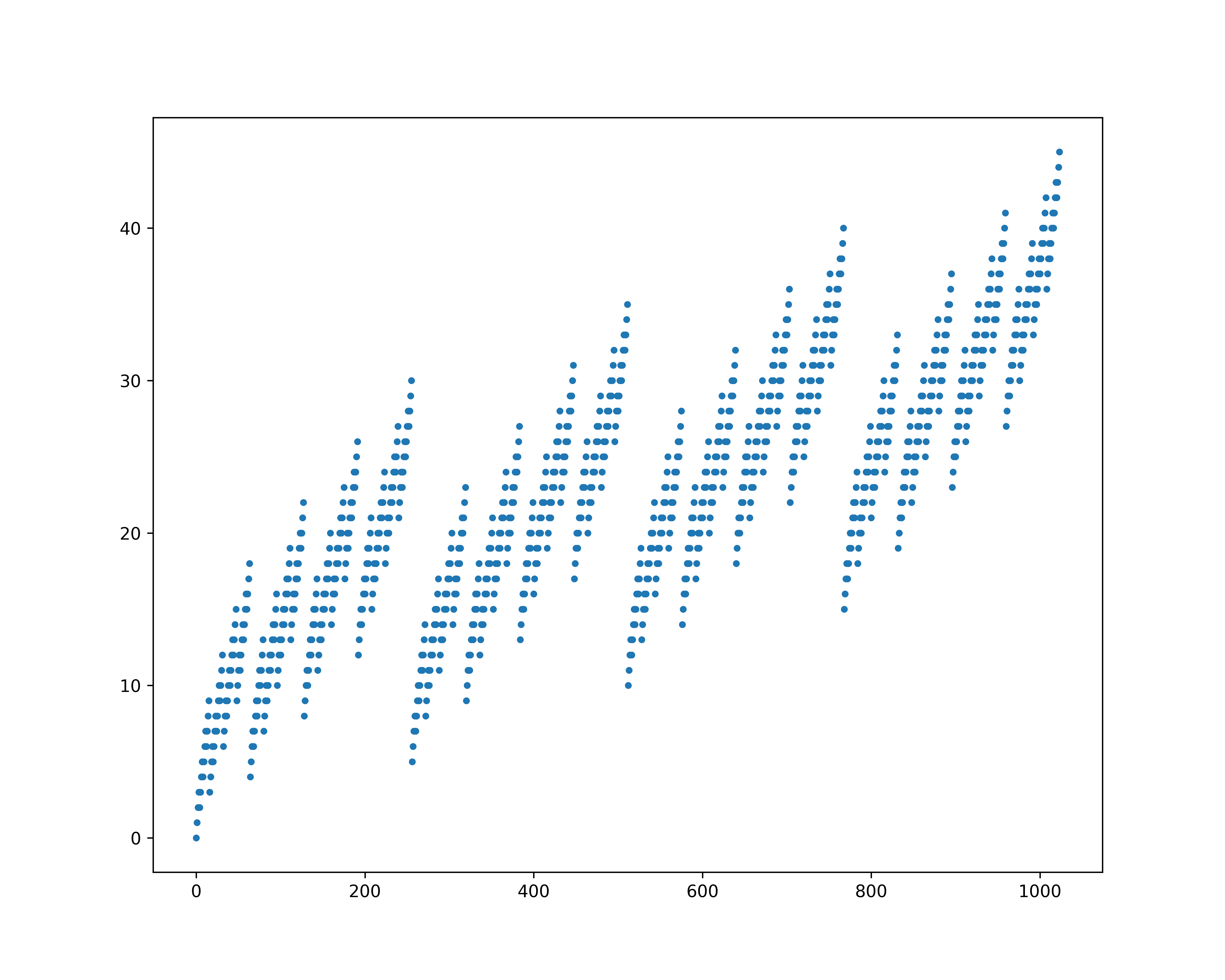}
  \caption{Mean values of the Gaussian emission probabilities used in the simulation experiments. The mean values are aligned according to the five consecutive outer state variations, indexed from $0$ to $1023$. The indexes are created by mapping $\{\A,\C,\G,\T\}$ to $\{0,1,2,3\}$ and viewing the five mapped numbers as a base-$4$ number system starting with the left-most position, e.g. `CAAAA' maps to $10000$ and is indexed $1$; `TTTTT' maps to $33333$ and is indexed $1023$. The mean values are generated by multiplying a kernel $\{1,2,3,4,5\}$ with the five mapped numbers as a vector.}
  \label{means}
\end{figure}

In the simulation experiment, we designed the outer states space to be $\mathbb{S}=\{\A,\C,\G,\T\}$ and the duration variable to have a Poisson distribution for $d\leq D$ and a geometric tail distribution with a fixed rate for $d>D$. The mean value for the Poisson distribution was set to $9$, $D$ was set to $16$, and the fixed rate for the exponential distribution was set to $0.5625$. The transition probability used in the base calling experiment, which was estimated using maximum likelihood estimation, or frequency count, from the reference genome, was utilized for generating $\hat{S}_{1:L}$. For each given reference length $L$ ranging from $500$ to $8000$, $200$ samples were generated by random sampling from the pre-obtained transition probability. Gaussian emission probabilities were used, where the observations were assumed to be a fifth-order Markov chain and were independent of duration variables, i.e., $P(X_t|\hat{S}_{l-4:l})\sim\mathcal{N}(m_i,\sigma)$ for $i\in\{1,...,4^5\}$. To facilitate comparisons of emission probabilities, we plotted the mean values of the Gaussian emission probabilities against all possible combinations of the five outer states in Figure \ref{means}. Standard deviations of the Gaussian emission probabilities were set to $2$ and $4$, respectively, to create scenarios of low and high noise ratios.

We used the PairwiseAligner function from Biopython packages \cite{biopython} to align the reference and decoded outer state sequence. 

\subsubsection{Results}
The mean values of the sequence accuracy for low and high emission noise scenarios are presented in Table \ref{acc_var2} and \ref{acc_var4}, respectively. A clear trend of increasing accuracy with increasing sequence length is observed. Tables \ref{len_var2} and \ref{len_var4} report the average number of matched outer states in the low and high noise scenarios. All methods naturally show a reduced performance with increased noise levels, as evidenced by decreased accuracy and fewer matched states.

The GMBS and the LFBS outperform the Viterbi algorithm in low and high noise scenarios. Specifically, the Viterbi gives the least number of matched states since it is prone to favor shorter lengths as they have a higher probability resulting from less frequent state transitions. In contrast, the modified BS algorithms with the marginalization step are less discriminative over the lengths of candidates. The GMBS generally achieves higher sequence accuracy and more matched outer bases than the LFBS. For both the GMBS and LFBS, a larger value of the beam width and the focus-length yields better performance, as expected. However, the performance gain becomes marginal as the computational cost grows. In particular, the improvement when increasing the beam width $W$ from $8$ to $64$ is most significant. Meanwhile, the GMBS with beam width $8$ displays a sign of degeneracy, which is comparable to the Viterbi and worse than the LFBS under the low noise scenario.

The decoding performances on the $k$-mer HPs are extracted from decoding results with reference lengths $1500$ and reported in Table \ref{simhomo2} and \ref{simhomo4} under the low and high noise scenarios, respectively. The results are clustered with respect to the lengths of the homopolymers, where the difficulty of correctly decoding rises as the length grows. In addition, since the data were created with a fifth-order Markov assumption, the HPs longer than $5$ are harder to decode. Similar to what is reflected in the decoding performances, the GMBS and the LFBS outperform the Viterbi in decoding HPs. A key point to consider is that the LFBS can not decode HPs longer than its focus-length $K$. The beams in the LFBS capturing various lengths of HPs were grouped together because they share the same last $K$ outer states, among which, later in the pruning step, the longer ones were pruned due to lower probability from more state transitions. Nevertheless, the LFBS gives the most numbers for the HPs of length less than $K$and the highest overall accuracy for all HPs of length at least 3.

The last Table \ref{timetable} presents the average decoding times per data for each method. For the GMBS, the number of logical threads issued per CUDA block \cite{cuda} equals the chosen beam width: $8$, $64$, $256$ and $512$. For the LFBS, the number of logical threads issued are $4^K$: $1024$ for $K=5$, $4096$ for $K=6$ and $16384$ for $K=7$. However, each physical thread on the GPU executes several logical threads because of the hardware limitation of at most 1024 threads per block and limits due to register memory on the GPU. Despite the exponential growth of the number of logical threads in the LFBS, the time cost of the LFBS scales almost linearly with respect to the focus-length $K$, similarly to the GMBS to the beam-width $W$. However, despite the theoretical advantages, the LFBS takes longer to finish in practice, most likely due to a memory I/O bottleneck on the GPU. This may, however, change on an implementation on dedicated hardware.
\begin{table}[!htp]
	\caption{Mean sequence accuracy with emission standard deviation $2$}
	\centering
\footnotesize
\footnotesize\begin{tabular}{l*{9}{c}}

\toprule
 \textbf{Ref Length}  & 1000 & 1500 & 2000 & 3000 & 4000 & 5000 & 6000 & 7000 & 8000 \\ 
 \hline
 GMBS(8)   & $0.9602$ & $0.9604$ & $0.9617$ & $0.9619$ & $0.9620$ & $0.9625$ & $0.9626$ & $0.9630$ & $ 0.9631$\\
 GMBS(64)  & $0.9695$ & $0.9695$ & $0.9705$ & $0.9701$ & $0.9707$ & $0.9711$ & $0.9710$ & $0.9713$ & $0.9714$\\  
 GMBS(256) & $0.9703$ & $0.9707$ & $0.9713$ & $0.9709$ & $0.9715$ & $0.9720$ & $0.9717$ & $0.9721$ & $0.9722$\\
 GMBS(512) & $\mathbf{0.9707}$ & $\mathbf{0.9709}$ & $\mathbf{0.9715}$ & $\mathbf{0.9711}$ & $\mathbf{0.9717}$ & $\mathbf{0.9722}$ & $\mathbf{0.9718}$ & $\mathbf{0.9722}$ & $\mathbf{0.9724}$\\
 LFBS(5)   & $0.9641$ & $0.9654$ & $0.9666$ & $0.9662$ & $0.9664$ & $0.9669$ & $0.9664$ & $0.9672$ & $0.9673$\\
 LFBS(6)   & $0.9660$ & $0.9671$ & $0.9683$ & $0.9678$ & $0.9684$ & $0.9689$ & $0.9682$ & $0.9691$ & $0.9691$\\
 LFBS(7)   & $0.9666$ & $0.9674$ & $0.9687$ & $0.9682$ & $0.9688$ & $0.9693$ & $0.9686$ & $0.9694$ & $0.9695$\\
 Viterbi   & $0.9592$ & $0.9608$ & $0.9615$ & $0.9618$ & $0.9627$ & $0.9630$ & $0.9629$ & $0.9628$ & $0.9631$\\
\toprule
\end{tabular}
\label{acc_var2}
\end{table}
\begin{table}[!htp]
	\caption{Mean sequence accuracy with emission standard deviation $4$}
	\centering
\footnotesize\begin{tabular}{l*{9}{c}}
\toprule
 \textbf{Ref Length}  & 1000 & 1500 & 2000 & 3000 & 4000 & 5000 & 6000 & 7000 & 8000 \\ 
 \hline
 GMBS(8)   & $0.7575$ & $0.7603$ & $0.7609$ & $0.7616$ & $0.7618$ & $0.7624$ & $0.7617$ & $0.7626$ & $0.7617$\\
 GMBS(64)  & $0.7740$ & $0.7749$ & $0.7755$ & $0.7758$ & $0.7762$ & $0.7771$ & $0.7767$ & $0.7774$ & $0.7759$\\
 GMBS(256) & $\mathbf{0.7753}$ & $\mathbf{0.7756}$ & $\mathbf{0.7762}$ & $\mathbf{0.7767}$ & $0.7767$ & $0.7778$ & $0.7771$ & $0.7780$ & $0.7767$\\
 GMBS(512) & $\mathbf{0.7753}$ & $0.7754$ & $\mathbf{0.7762}$ & $0.7763$ & $\mathbf{0.7769}$ & $\mathbf{0.7780}$ & $\mathbf{0.7772}$ & $\mathbf{0.7781}$ & $\mathbf{0.7768}$\\
 LFBS(5)   & $0.7404$ & $0.7410$ & $0.7413$ & $0.7431$ & $0.7426$ & $0.7436$ & $0.7430$ & $0.7435$ & $0.7426$\\
 LFBS(6)   & $0.7436$ & $0.7443$ & $0.7440$ & $0.7457$ & $0.7459$ & $0.7458$ & $0.7456$ & $0.7466$ & $0.7457$\\
 LFBS(7)   & $0.7438$ & $0.7453$ & $0.7439$ & $0.7462$ & $0.7464$ & $0.7467$ & $0.7468$ & $0.7473$ & $0.7466$\\
 Viterbi   & $0.7106$ & $0.7117$ & $0.7120$ & $0.7122$ & $0.7129$ & $0.7131$ & $0.7127$ & $0.7133$ & $0.7134$\\
\toprule
\end{tabular}
\label{acc_var4}
\end{table}
\begin{table}[!htp]
	\caption{Average number of matched outer states with emission standard deviation $2$}
	\centering
\footnotesize\begin{tabular}{l*{9}{c}}
\toprule
 \textbf{Ref Length}  & 1000 & 1500 & 2000 & 3000 & 4000 & 5000 & 6000 & 7000 & 8000 \\ 
 \hline
 GMBS(8)  &  $982$& $1473$& $1965$& $2948$& $3932$& $4916$& $5901$& $6883$& $7869$\\
 GMBS(64) &  $987$& $1480$& $1971$& $2962$& $3947$& $4933$& $5925$& $6916$& $7904$\\  
 GMBS(256)&  $\mathbf{988}$& $1481$& $1972$& $2962$& $3948$& $4934$& $5925$& $6916$& $7906$\\
 GMBS(512)&  $\mathbf{988}$& $\mathbf{1483}$& $\mathbf{1973}$& $\mathbf{2963}$& $\mathbf{3950}$& $\mathbf{4937}$& $\mathbf{5929}$& $\mathbf{6919}$& $\mathbf{7910}$\\
 LFBS(5)  &  $987$& $1479$& $1970$& $2956$& $3943$& $4930$& $5918$& $6909$& $7893$ \\
 LFBS(6)  &  $987$& $1481$& $1971$& $2959$& $3950$& $4935$& $5926$& $6916$& $7906$\\
 LFBS(7)  &  $\mathbf{988}$& $1482$& $1972$& $2958$& $3950$& $4935$& $\mathbf{5929}$& $6916$& $7907$\\
 Viterbi  &  $974$& $1462$& $1944$& $2923$& $3903$& $4871$& $5851$& $6828$& $7805$\\
 \toprule
\end{tabular}
\label{len_var2}
\end{table}
\begin{table}[!htp]
	\caption{Average number of matched outer states with emission standard deviation $4$}
	\centering
\footnotesize\begin{tabular}{l*{9}{c}}
\toprule
 \textbf{Ref Length}  & 1000 & 1500 & 2000 & 3000 & 4000 & 5000 & 6000 & 7000 & 8000\\ 
 \hline
 GMBS(8)  &  $859$& $1291$& $1722$& $2585$& $3447$& $4312$& $5172$& $6037$& $6895$\\
 GMBS(64) &  $\mathbf{882}$& $1316$& $1751$& $2614$& $3509$& $4384$& $5232$& $6137$& $7004$\\  
 GMBS(256)&  $\mathbf{882}$& $1315$& $1751$& $2616$& $\mathbf{3517}$& $4397$& $5240$& $6144$& $7011$\\
 GMBS(512)&  $\mathbf{882}$& $\mathbf{1318}$& $\mathbf{1753}$& $\mathbf{2618}$& $\mathbf{3517}$& $\mathbf{4400}$& $\mathbf{5247}$& $\mathbf{6151}$& $\mathbf{7024}$\\
 LFBS(5)  &  $876$& $1311$& $1746$& $2602$& $3494$& $4376$& $5224$& $6111$& $6982$\\
 LFBS(6)  &  $880$& $1314$& $1748$& $2610$& $3508$& $4384$& $5228$& $6128$& $7004$\\
 LFBS(7)  &  $881$& $1315$& $1745$& $2614$& $3506$& $4382$& $5235$& $6135$& $6992$\\
 Viterbi  &  $810$& $1218$& $1630$& $2424$& $3253$& $4083$& $4857$& $5678$& $6463$\\
 \toprule
\end{tabular}
\label{len_var4}
\end{table}

\begin{table}[!htp]\caption{Decoding performances on homopolymers with ref length $1500$ and emission standard deviation $2$}\centering
\footnotesize\begin{tabular}{lcccccccr}
\toprule
\textbf{Methods}&\textbf{3-mer} &\textbf{4-mer} &\textbf{5-mer} &\textbf{6-mer} &\textbf{7-mer} &\textbf{8-mer}&\textbf{9-mer}&Accuracy \\\hline
GMBS(8)  &$9799/10482$&$2653/2884$&$816/908$&$197/254$&$38/54$&$\mathbf{7}/11$&$\mathbf{4}/8$&$0.9256$\\
GMBS(64) &$9960/10499$&$2690/2884$&$824/909$&$199/254$&$38/54$&$\mathbf{7}/11$&$\mathbf{4}/8$&$0.9387$\\
GMBS(256)&$9960/10505$&$2693/2885$&$824/909$&$199/254$&$\mathbf{39}/54$&$\mathbf{7}/11$&$\mathbf{4}/8$&$0.9385$\\
GMBS(512)&$9965/10505$&$2695/2885$&$824/909$&$199/254$&$\mathbf{39}/54$&$\mathbf{7}/11$&$\mathbf{4}/8$&$0.9390$\\
LFBS(5)  &$9996/10497$&$2713/2888$&$\mathbf{825}/909$&$0/254$&$0/53$&$0/11$&$0/8$&$0.9258$\\
LFBS(6)  &$10008/10500$&$2723/2888$&$818/909$&$197/255$&$0/53$&$0/11$&$0/8$&$0.9400$\\
LFBS(7)  &$\mathbf{10016}/10500$&$\mathbf{2724}/2887$&$818/909$&$198/254$&$37/54$&$0/11$&$0/8$&$\mathbf{0.9433}$\\
Viterbi  &$9672/10493$&$2640/2885$&$776/909$&$139/254$&$21/54$&$3/11$&$1/8$&$0.9069$\\ \toprule
\end{tabular}\\{*Results are displayed as numbers of `correctly-decoded/total homopolymers'.}
\label{simhomo2}
\end{table}

\begin{table}[!htp]\caption{Decoding performances on homopolymers with ref length $1500$ and emission standard deviation $4$}\centering
\footnotesize\begin{tabular}{lcccccccr}
\toprule
\textbf{Methods}&\textbf{3-mer} &\textbf{4-mer} &\textbf{5-mer} &\textbf{6-mer} &\textbf{7-mer} &\textbf{8-mer}&\textbf{9-mer}&Accuracy \\\hline
GMBS(8)  &$6665/10381$&$1750/2822$&$572/887$&$144/248$&$31/70$&$8/17$&$3/4$&$0.6357$\\
GMBS(64) &$6920/10398$&$1853/2822$&$605/888$&$146/247$&$\mathbf{33}/70$&$\mathbf{10}/17$&$\mathbf{4}/4$&$0.6625$\\
GMBS(256)&$6900/10397$&$1858/2819$&$\mathbf{612}/890$&$146/247$&$\mathbf{33}/70$&$9/17$&$\mathbf{4}/4$&$0.6620$\\
GMBS(512)&$6910/10406$&$1865/2822$&$\mathbf{612}/890$&$\mathbf{147}/247$&$\mathbf{33}/70$&$\mathbf{10}/17$&$\mathbf{4}/4$&$0.6628$\\
LFBS(5)  &$6907/10353$&$1875/2817$&$559/887$&$0/247$&$0/70$&$0/17$&$0/4$&$0.6489$\\
LFBS(6)  &$6947/10359$&$1912/2815$&$579/886$&$126/249$&$0/69$&$0/17$&$0/4$&$0.6642$\\
LFBS(7)  &$\mathbf{7000}/10371$&$\mathbf{1924}/2817$&$595/888$&$138/248$&$36/70$&$0/17$&$0/4$&$\mathbf{0.6724}$\\
Viterbi  &$6075/10216$&$1694/2788$&$514/877$&$108/244$&$16/70$&$4/17$&$1/4$&$0.5917$\\ \toprule
\end{tabular}\\{*Results are displayed as numbers of `correctly-decoded/total homopolymers'.}
\label{simhomo4}
\end{table}

\begin{table}[!htp]
	\caption{Mean time cost (seconds)}
	\centering
\footnotesize\begin{tabular}{l*{10}{c}}
\toprule
 \textbf{Ref Length}  & 1000 & 1500 & 2000 & 3000 & 4000 & 5000 & 6000 & 7000 & 8000 \\ 
 \hline
 GMBS(8)   & $0.4120$ & $0.6158$ & $0.8043$ & $1.1726$ & $1.5888$ & $1.9413$ & $2.3157$ & $2.7226$ & $3.1338$\\
 GMBS(64)  & $0.9067$ & $1.3523$ & $1.7986$ & $2.6936$ & $3.5958$ & $4.4971$ & $5.4072$ & $6.2591$ & $7.2820$\\  
 GMBS(256) & $2.9821$ & $4.4884$ & $5.9852$ & $8.9905$ &$11.9829$ &$15.0802$ &$18.0231$ &$21.1296$ &$24.0776$\\
 GMBS(512) & $6.3481$ & $9.6344$ & $12.8240$&$19.2897$ &$25.5928$ &$32.1701$ &$38.5581$ &$44.8304$ &$51.4094$\\
 LFBS(5)   & $0.7342$ & $1.105$  & $1.4660$ & $2.2106$ & $2.9409$ & $3.6768$ & $4.4093$ & $5.1546$ & $5.8806$\\
 LFBS(6)   & $2.7459$ & $4.1234$ & $5.5016$ & $8.2658$ & $11.0160$&$13.8055$ &$16.5822$ &$19.3067$ &$22.0654$\\
 LFBS(7)   & $11.1187$& $16.5055$& $22.0554$&$33.0467$ & $43.7554$&$53.9086$ &$70.2276$ &$77.7335$ &$90.3488$\\
 \toprule
\end{tabular}
\label{timetable}
\end{table}

\subsection{Base Calling Experiment and Results}
In the second experiment, we used the Lokatt model and data batch $2$ from \cite{Lokatt}. The dataset comprises raw reads (ion current measurements) generated by ONT MinION devices with the $1$D protocol\cite{Jain2016}, and their corresponding Ecoli genome (nucleotide bases) as the reference sequence. The hybrid Lokatt model includes an EDHMM that employs the output of an NN in place of the emission probabilities. The pre-trained NN takes a raw read $X_{1:T}$ as input and generates the sequence of likelihoods for $5$ consecutive bases, referred to as $5$-mers. The duration variable was set to follow a log-logistic distribution with an geometric tail, which was estimated with respect to each raw read based on the number of changes above a threshold level in the current amplitudes.

We used the software Minimap2 \cite{Minimap2} to map the decoded outer states to the reference genome. The reads were selected into two categories, \textit{short reads} and \textit{long reads}, corresponding to lengths of $1000-2000$ and $4000-5000$ bases, respectively. To account for the random noise in the dataset, we evaluated the identity and number of matched bases in terms of both their means and medians.

\subsubsection{Results}
The results of the decoding experiment with short reads and long reads are reported in Table \ref{dna1} and \ref{dna2}, respectively. We observe a slight decrease in the mean identity scores of all methods as the length increases, contrary to what was observed in the simulation experiment. This is likely due to the systematic errors in the devices, such as the ratcheting motor protein controlling the sequencing speed, leading to slightly lower quality measurements in long reads. 

Although the modified BS algorithms demonstrated superior performance over the Viterbi algorithm in our simulation experiments, these benefits did not always remain when applied to a real-world base calling dataset. In particular, the Viterbi algorithm's performance is comparable to the GMBS and the LFBS, even at their maximum configurations. The outstanding behavior of the Viterbi algorithm has been observed in previous reports based on real datasets, such as those in \cite{durationviterbi,Chiron,Bonito}. A possible explanation is the Lokatt model does not precisely match the underlying data-generating process of the base calling dataset. This mismatch may cause the additional marginalization step to contribute to incorrect predictions of the states. Nevertheless, all methods' mean and median identity scores exhibited only minor variations. The LFBS with focus-length $6$ and $7$ decoded the highest number of bases with a leading identity score in both cases.

The decoding performances of the homopolymers are presented in Table \ref{homo1} and \ref{homo2}. The long HPs in the base calling dataset are harder to decode, resulting in all methods having lower performance than in the simulation experiments. In contrast to the identity scores, the marginalized BS algorithms outperform the Viterbi in decoding HPs of various lengths. The Viterbi typically has the least number of HPs correct and fails to decode homopolymers longer than $5$. The GMBS with $512$ beams shows superior performance regarding the number of correctly called HPs and overall accuracy. 
The LFBS, which can not decode HPs longer than the chosen $K$, caught up with the GMBS for decoding long HPs as the values of $K$ increased.

\begin{table}[!htp]\caption{Decoding performances with short reads}\centering
\footnotesize\begin{tabular}{lcccc}
\toprule
\textbf{Methods}&\textbf{Identity} & \textbf{Bases per read} & \textbf{Reads} &\textbf{Matched Bases}\\\hline
GMBS(8)  & $0.9141/0.9269$ & $1669/1805$ & $455$ & $759781$\\
GMBS(64) & $0.9152/0.9303$ & $1672/1816$ & $458$ & $765812$\\
GMBS(256)& $0.9161/0.9307$ & $1679/\mathbf{1818}$ & $456$ & $765411$\\
GMBS(512)& $\mathbf{0.9167}/0.9309$ & $\mathbf{1681}/1817$ & $455$ & $765048$\\
LFBS(5)  & $0.9154/\mathbf{0.9312}$ & $1671/1811$ & $\mathbf{459}$ & $766902$\\
LFBS(6)  & $0.9156/0.9311$ & $1671/1811$ & $\mathbf{459}$ & $767012$\\
LFBS(7)  & $0.9159/\mathbf{0.9312}$ & $1675/1812$ & $458$ & $\mathbf{767060}$\\
Viterbi  & $0.9158/\mathbf{0.9312}$ & $1671/1810$ & $458$ & $765634$\\\toprule
\end{tabular}\\{*Identity and the number of matched bases per read are listed as `mean/median'.}
\label{dna1}
\end{table}
\begin{table}[!htp]\caption{Decoding performances with long reads}\centering
\footnotesize\begin{tabular}{lcccc}
\toprule
\textbf{Methods}&\textbf{Identity} & \textbf{Bases per read} & \textbf{Reads} &\textbf{Matched Bases}\\\hline
GMBS(8)  & $0.9102/0.9277$ & $4193/4504$ & $544$ & $2281135$\\
GMBS(64) & $0.9111/0.9298$ & $4206/4510$ & $546$ & $2296959$\\
GMBS(256)& $0.9115/0.9305$ & $4215/4510$ & $546$ & $2301715$\\
GMBS(512)& $0.9112/0.9299$ & $4217/\mathbf{4514}$ & $546$ & $2302800$\\
LFBS(5)  & $0.9121/0.9311$ & $4210/4513$ & $\mathbf{547}$ & $2303138$\\
LFBS(6)  & $0.9123/0.9315$ & $\mathbf{4223}/4512$ & $546$ & $\mathbf{2306013}$\\
LFBS(7)  & $\mathbf{0.9124}/0.9315$ & $\mathbf{4223}/4513$ & $546$ & $2305796$\\
Viterbi  & $0.9122/\mathbf{0.9318}$ & $4216/4509$ & $546$ & $2302418$\\\toprule
\end{tabular}\\{*Identity and the number of matched bases per read are listed as `mean/median'.}
\label{dna2}
\end{table}

\begin{table}[!htp]\caption{Decoding performances on homopolymers with short reads}\centering
\footnotesize\begin{tabular}{lccccccr}
\toprule
\textbf{Methods}&\textbf{3-mer} &\textbf{4-mer} &\textbf{5-mer} &\textbf{6-mer} &\textbf{7-mer} &\textbf{8-me}&Accuracy \\\hline
GMBS(8)  &$3248/3966$&$\mathbf{717}/1047$&$145/320$&$5/91$&$0/23$&$0/10$&$0.7542$\\
GMBS(64) &$3286/3998$&$716/1054$&$146/323$&$\mathbf{6}/91$&$0/23$&$0/10$&$0.7555$\\
GMBS(256)&$3288/4005$&$\mathbf{717}/1054$&$\mathbf{148}/323$&$\mathbf{6}/91$&$0/23$&$0/10$&$0.7555$\\
GMBS(512)&$3276/3975$&$715/1049$&$\mathbf{148}/321$&$\mathbf{6}/91$&$0/23$&$0/10$&$\mathbf{0.7580}$\\
LFBS(5)  &$3290/4008$&$701/1054$&$132/323$&$0/91$&$0/23$&$0/10$&$0.7485$\\
LFBS(6)  &$\mathbf{3292}/4008$&$710/1054$&$138/323$&$3/91$&$0/23$&$0/10$&$0.7522$\\
LFBS(7)  &$3291/4008$&$711/1054$&$139/323$&$5/91$&$0/23$&$0/10$&$0.7527$\\
Viterbi  &$3271/3977$&$699/1049$&$133/320$&$0/91$&$0/23$&$0/10$&$0.7502$\\ \toprule
\end{tabular}\\{*Results are displayed as numbers of `correctly-decoded/total homopolymers'.}
\label{homo1}
\end{table}
\begin{table}[!htp]\caption{Decoding performances on homopolymers with long reads}\centering
\footnotesize\begin{tabular}{lcccccccr}
\toprule
\textbf{Methods}&\textbf{3-mer} &\textbf{4-mer} &\textbf{5-mer} &\textbf{6-mer} &\textbf{7-mer} &\textbf{8-mer}&\textbf{9-mer}&Accuracy \\\hline
GMBS(8)  &$13293/16022$&$3042/4224$&$\mathbf{652}/1338$&$\mathbf{23}/369$&$1/72$&$0/23$&$0/2$&$0.7715$\\
GMBS(64) &$13363/16046$&$3044/4225$&$649/1337$&$\mathbf{23}/369$&$\mathbf{2}/72$&$0/23$&$0/2$&$0.7738$\\
GMBS(256)&$13363/16027$&$3045/4225$&$647/1336$&$\mathbf{23}/368$&$\mathbf{2}/73$&$0/23$&$0/2$&$0.7745$\\
GMBS(512)&$\mathbf{13380}/16037$&$\mathbf{3047}/4231$&$650/1338$&$\mathbf{23}/369$&$\mathbf{2}/73$&$0/23$&$0/2$&$\mathbf{0.7748}$\\
LFBS(5)  &$13370/16037$&$2994/4226$&$601/1335$&$0/368$&$0/73$&$0/23$&$0/2$&$0.7689$\\
LFBS(6)  &$13375/16037$&$3026/4225$&$630/1336$&$13/368$&$0/73$&$0/23$&$0/2$&$0.7725$\\
LFBS(7)  &$13371/16037$&$3027/4225$&$635/1336$&$21/368$&$2/73$&$0/23$&$0/2$&$0.7731$\\
Viterbi  &$13328/16029$&$2989/4226$&$585/1335$&$0/368$&$0/73$&$0/23$&$0/2$&$0.7664$\\ \toprule
\end{tabular}\\{*Results are displayed as numbers of `correctly-decoded/total homopolymers'.}
\label{homo2}
\end{table}
\section{Conclusion}
In this paper, we have proposed the GMBS and the LFBS algorithms to approximate the solution for decoding the outer state sequences on a given HHMM. Our investigation into the performance of the two marginalized BS algorithms in simulated and real-world datasets has shown that they can outperform the Viterbi in certain scenarios. However, the advantages are insignificant when applied to real-world datasets, despite the Viterbi achieving fewer matched bases. The GMBS and the LFBS did not provide greater identity scores but generally matched more bases.

We also observed that the performance of different algorithms varies depending on the type of sequences being decoded, such as homopolymers of different lengths. The Viterbi has performed worse in terms of accuracy and numbers than the GMBS and the LFBS in decoding homopolymers. Moreover, the computational cost of the GMBS and the LFBS in a parallel implementation increased linearly with their parameter value, though are still more costly than the Viterbi. When choosing the decoding algorithm, one must therefore consider the trade-off between accuracy and speed.

These findings provide insights into the selection and optimization of BS algorithms in base calling applications and may guide future research in this area. However, our study has some limitations. Our analysis was based on a specific base calling dataset and may not generalize to other datasets with different characteristics, such as natural language datasets. Future work could expand the scope of the study by testing more algorithms and evaluating their performance on different datasets. Additionally, it would be interesting to explore the effect of parameter tuning on the performance of the BS algorithms, as well as the potential benefits of combining different algorithms or developing novel hybrid methods.

Last but not least, the two BS algorithms proposed in this work are also, in principle, applicable to graphic models without explicit Markov properties, such as the Connectionist Temporal Classification (CTC) graphs \cite{CTC}. We will, however, leave the exploration of this to future work. 

%% file: MGBS_algo.tex
\begin{algorithm}\footnotesize
  \caption{The GMBS algorithm}\label{GMBS}
  \begin{algorithmic}[1]
    \myState{\textbf{Input:} $X_{1:T}$, HHMM, $W$}
    \myState{\textbf{Output:} the sequence $\hat{S}_{1:L}$}
    
        \myState{Initialise with priors $P(\hat{S})$}
        \myState{\textbf{At} $\mathbf{t=1}$}
        \myState{$1.$ Create leaf beams $(\hat{S}_1,X_1)$ for all $\hat{S}_1\in\mathbb{S}$ and}
        \myState{2. assign each a score $P(\hat{S}_1,X_1)=P(X_1|\hat{S}_1)P(\hat{S}_1)$.}
        \myState{$3.$ Sort and keep the top-$W$ highest-scored leaf beams as the beam.}
        \myState{$4.$ Assign a unique ID with timestamp $t$ to each beam.}
        \For{$\mathbf{t=2,...,T}$}
            \Procedure {Expanding}{}    
            \For{Every beam $(\hat{S}_{1:l^w_{t-1}},X_{1:t-1})$ (in parallel) :}
                \myState{$1.$ Expand $|\mathbb{S}|$ vanilla leaf beams $(\hat{S}_{1:(l^w_{t-1}+1)},X_{1:t})$ and compute their scores, and}
                \myState{$2.$ assign each a unique ID with timestamp $t$, with the ID of the original beam as their parent ID.}
                \myState{$3.$ Expand a vanilla leaf beam $(\hat{S}_{1:l^w_{t-1}},X_{1:t})$, compute its score, and}
                \myState{$4.$ maintain the same ID and the same parent ID of the original beam.}
            \EndFor
          \EndProcedure
            \Procedure {Marginalizing}{}    
                \myState{1. Sort the parent ID of all vanilla leaf beams, and}
                \myState{2. for these sharing the same parent ID, check if they also share the same last state}
                \myState{3. Merge the redundant vanilla leaf beams as described in Chapter \ref{IDs}.}
                \myState{4. Promote the non-redundant vanilla leaf beams as leaf beams.}
            \EndProcedure
            \Procedure {Pruning}{}        
                \myState{$1.$ Sort remaining leaf beams by their scores.}
                \myState{$2.$ Promote the top-$W$ leaf beams to the beams $(\hat{S}_{1:l^w_{t}},X_{1:t})$.}
            \EndProcedure
        \EndFor
  
    \Procedure{Backwards-tracking}{}
    \myState{$1.$ Select the highest-scored beam $(\hat{S}_{1:l^w_{T}},X_{1:T})=\argmax\Bar{\chi}_T(\hat{S}_{1:l^w_{T}})$.}
    \myState{$2.$ Back trace to the ancestor beam pinpointed by the parent ID of the selected beam.}
    \myState{$3.$ Concatenating the recorded last outer states in the parent beam.}
    \myState{$4.$ If the timestamp in the parent ID is $0$, end the procedure, otherwise,}
    \myState{5. make the parent beam the selected beam and return to 2.}
    \EndProcedure
  \end{algorithmic}
  \end{algorithm}

%% file: LFBS_algo.tex
\begin{algorithm}\footnotesize
  \caption{LFBS algorithm}\label{LFBS}
  \begin{algorithmic}[1]
    \myState{\textbf{Input:} $X_{1:T}$, HHMM, $K$, $M=|\mathbb{S}|^K$}
    \myState{\textbf{Output:} $\hat{S}_{1:L}$}
    
        \myState{Initialise with priors $P(A_m)$ for $m=1,...,M$}
        \myState{\textbf{At} $\mathbf{t=1}$}
        \myState{$1.$ Create $M$ beams $(\hat{S}_{1:K},X_1)$ for all $\hat{S}_{1:K}=A_m\in\mathbb{S}_{1:K}$ and}
        \myState{2. assign $m$th beam a score $P(A_m,X_1)=P(X_1|A_m)P(A_m)$.}
        \myState{$3.$ Assign a unique ID with timestamp $t$ to each beam.}
        \For{$\mathbf{t=2,...,T}$}
            \Procedure {Expanding}{}    
            \For{ the $m$th beam $(\hat{S}_{1:l^m_{t-1}},X_{1:t-1})$ (in parallel) :}
                \myState{$1.$ Expand $|\mathbb{S}|$ vanilla leaf beams $(\hat{S}_{1:(l^m_{t-1}+1)},X_{1:t})$ and compute their scores, and}
                \myState{$2.$ assign each a unique ID with timestamp $t$, with the ID of the original beam as their parent ID.}
                \myState{$3.$ Expand a continuation vanilla leaf beam $(\hat{S}_{1:l^m_{t-1}},X_{1:t})$, compute its score, and}
                \myState{$4.$ maintain the same ID and the same parent ID of the original beam.}
            \EndFor
          \EndProcedure
          \For{the $m$th group of vanilla leaf beams with ending state sequence $A_m$ (in parallel)}
            \Procedure {Marginalizing}{}    
            
                \myState{1. Sort the parent ID of all vanilla leaf beams, and}
                \myState{2. for leaf beams sharing the same parent ID, check if they also share the same last state}
                \myState{3. Merge the redundant vanilla leaf beams as described in Chapter \ref{IDs}.}
                \myState{4. Promote the non-redundant vanilla leaf beams as leaf beams.}
                  \EndProcedure
                  \Procedure{Pruning}{}
                        \myState{$1.$ Promote the highest-scored leaf beam as the new beam $(\hat{S}_{1:l^m_t},X_{1:t})$.}
                    \EndProcedure
            \EndFor
    \EndFor
    \Procedure{Backwards-tracking}{}
    \myState{$1.$ Select the highest-scored beam $(\hat{S}_{1:l^m_{T}},X_{1:T})=\argmax\Bar{\chi}_T(\hat{S}_{1:l^m_{T}})$.}
    \myState{$2.$ Back trace to the ancestor beam pinpointed by the parent ID of the selected beam.}
    \myState{$3.$ Concatenating the recorded last outer states in the parent beam.}
    \myState{$4.$ If the timestamp in the parent ID is $0$, end the procedure, otherwise,}
    \myState{5. make the parent beam the selected beam and return to 2.}
    \EndProcedure
  \end{algorithmic}
  \end{algorithm}

%% file: appendix.tex
\def\snr{\mathrm{SNR}}
\def\nt{n_\mathrm{t}}
\def\nr{n_\mathrm{r}}
\def\cnormal{\mathcal{N}_\complex}
\def\fro{_\mathrm{F}}
\def\kron{\otimes}

\def\LB{_{\mathrm{L}}}
\def\UB{_{\mathrm{U}}}

\section*{Appendix A}

We will derive \eqref{twosum} by induction using the two-level HHMM described in chapter \ref{chap2}, where we divided the inner state space into the continuation set $\mathbb{S}^2_\text{c}$, and the end set $\mathbb{S}^2_\text{e}$. We let $\mathbb{B}_t(\hat{S}_{1:l})$ be the joint time-aligned sequences $S_{1:t} = (S^1_{1:t}, S^2_{1:t})$ that map to a sequence-aligned outer state sequence $\hat{S}_{1:l}$. The objective of the decoding problem essentially is to, given an observation sequence $X_{1:t}$, compute or approximate the metric
$$
P(X_{1:t}, \hat{S}_{1:l}) = \sum_{S_{1:t} \in \mathbb{B}_t(\hat{S}_{1:l})} P(X_{1:t},S_{1:t})
$$
for all candidate sequences $\hat{S}_{1:L}$ and select the maximum of these sequences.

Let $\chi_t(\hat{S}_{1:l},i)$ denotes the probability of the sequence-aligned states $\hat{S}_{1:l}$ with a specific current inner state $S^2_t=i$ with $i\in \mathbb{S}^2$ such that
\begin{equation}
\chi_t(\hat{S}_{1:l},i)=\sum_{S_{1:t}\in\mathbb{B}_t(\hat{S}_{1:l}),S^2_t=i}P(S_{1:t},X_{1:t}) \, .
\label{setB}
\end{equation}
Further, let $\mathbb{B}^\text{e}_{t} (\hat{S}_{1:l}) $ be given by
$$
\mathbb{B}^\text{e}_{t} (\hat{S}_{1:l}) = \big\{S_{1:t}  \, | \, S_{1:{t-1}}  \in \mathbb{B}_{t-1}(\hat{S}_{1:l-1}), \, S_{t-1}=(\hat{S}_{l-1},j) \big\} \, 
$$
if $j \in \mathbb{S}^2_\text{e}$, and
$$
\mathbb{B}^\text{c}_{t} (\hat{S}_{1:l}) = \big\{S_{1:t}  \, | \, S_{1:t-1}  \in \mathbb{B}_{t-1}(\hat{S}_{1:l}) , \, S_{t-1}=(\hat{S}_l,k)\big\}
$$
if $k \in \mathbb{S}^2_\text{c}$. By construction, we have $
\mathbb{B}_t(\hat{S}_{1:l}) = \mathbb{B}^\text{e}_{t} (\hat{S}_{1:l})\cup \mathbb{B}^\text{c}_{t} (\hat{S}_{1:l})
$ while  $\mathbb{B}^\text{e}_{t} (\hat{S}_{1:l}) \cap \mathbb{B}^\text{c}_{t} (\hat{S}_{1:l})= \emptyset$, which implies that
$$
\chi_t(\hat{S}_{1:l}, i) = \sum_{S_{1:t}\in\mathbb{B}^\text{e}_t(\hat{S}_{1:l}),S^2_t=i}P(S_{1:t},X_{1:t})+\sum_{S_{1:t}\in\mathbb{B}^\text{c}_t(\hat{S}_{1:l}),S^2_t=i}P(S_{1:t},X_{1:t})
$$
Note now that for any $S_{1:t} \in \mathrm{B}_{t}(\hat{S}_{1:l})$, we can obtain
\begin{align*}
& P(S_{1:t}, X_{1:t}) \\
=& P(S_t, X_t,  S_{1:t-1}, X_{1:t-1}) \\
=& P(S_t, X_t | S_{1:t-1}, X_{1:t-1}) P(S_{1:t-1}, X_{1:t-1}) \, ,
\end{align*}
and use the HHMM assumption to conclude that
$$
P(S_t, X_t | S_{1:t-1}, X_{1:t-1}) = P(S_t, X_t | S_{t-1}).
$$ 
 This implies that
\begin{align*}
\chi_t(\hat{S}_{1:l}, i) & = \sum_{S_{1:t}\in\mathbb{B}^\text{e}_t(\hat{S}_{1:l}),S^2_t=i}P(S_{1:t},X_{1:t})+\sum_{S_{1:t}\in\mathbb{B}^\text{c}_t(\hat{S}_{1:l}),S^2_t=i}P(S_{1:t},X_{1:t})\\
&= \sum_{S_{1:t}\in\mathbb{B}^\text{e}_t(\hat{S}_{1:l}),S^2_t=i}P(S_t,X_t|S_{t-1})P(X_{1:t-1}, S_{1:t-1}) \\
& \qquad \qquad + \sum_{S_{1:t}\in\mathbb{B}^\text{c}_t(\hat{S}_{1:l}),S^2_t=i}P(S_t,X_t|S_{t-1})P(X_{1:t-1}, S_{1:t-1})\\
& =\sum_{j\in\mathbb{S}^2_\text{e}}P(S_t,X_t|S_{t-1}=(\hat{S}_{l-1},j))\sum_{S_{1:t-1}\in\mathbb{B}_{t-1}(\hat{S}_{1:l-1}),S^2_{t-1}=j}P(S_{1:t-1},X_{1:t-1})\\
&\qquad \qquad + \sum_{k\in\mathbb{S}^2_\text{c}}P(S_t,X_t|S_{t-1}=(\hat{S}_{l},k))\sum_{S_{1:t-1}\in\mathbb{B}_{t-1}(\hat{S}_{1:l}),S^2_{t-1}=k}P(S_{1:t-1},X_{1:t-1})\\
&= \sum_{j \in \mathbb{S}^2_\text{e}}  P\big(S_t,X_t| S_{t-1}=(\hat{S}_{l-1},j)\big) \chi_{t-1}(\hat{S}_{1:l-1}, j) \\
& \qquad + \sum_{k \in \mathbb{S}^2_\text{c}} P\big(S_t,X_t| S_{t-1}=(\hat{S}_{l},k)\big) \chi_{t-1}(\hat{S}_{1:l}, k) \, .
\end{align*}
In other words, one can recursively compute $\chi_t(\hat{S}_{1:l},i)$ for any sequence $\hat{S}_{1:l}$ and the inner state $i$ from the values of $\chi_{t-1}$. \comment{We can further simplify the model if we assume that
$$
P(X_t, S_t | S_{t-1}) = P(X_t, | S_t, \hat{S}_k, S^2_{t-1}) P(S_t | \hat{S}_k, S^2_{t-1}) = P(X_t, | S_t) P(S_t | \hat{S}_k, S^2_{t-1}) 
$$
in which case we get
$$
\chi_t(\hat{S}_{1:k}, j) =  P(X_t, | S_t) \left( \sum_{i \in \mathcal{E}} P(S_t | \hat{S}_k, S^2_{t-1}) \chi_{t-1}(\hat{S}_{1:k-1}, i) + \sum_{i \in \mathcal{S}} P(S_t | \hat{S}_k, S^2_{t-1}) \chi_{t-1}(\hat{S}_{1:k}, i) \right) \, .
$$
Finally, note that
$$
P(S_t | \hat{S}_k, S^2_{t-1}) = P(S_t | \hat{S}_k, S^2_{t-1} = i)
$$
when $i \in \mathcal{E}$ and 
$$
P(S_t | \hat{S}_k, S^2_{t-1}) = P(S_t | \hat{S}_k, S^2_{t-1} = i)
$$
and $i \in \mathcal{S}$ have simple expressions given by the HHMM transitions probabilities.}

%% file: main.bbl
\begin{thebibliography}{10}

\bibitem{Kamaric1999}
S{\o}ren Riis.
\newblock {\em Hidden {M}arkov models and neural networks for speech
  recognition}.
\newblock Technical University of Denmark [Department of Mathematical
  Modeling], 1998.

\bibitem{speech}
Mark Gales and Steve Young.
\newblock {\em Application of Hidden {M}arkov Models in Speech Recognition}.
\newblock Now Foundations and Trends, 2008.

\bibitem{HHMMhealthcare}
Ya-Ti Peng, Ching-Yung Lin, Ming-Ting Sun, and Kun-Cheng Tsai.
\newblock Healthcare audio event classification using hidden {M}arkov models
  and hierarchical hidden {M}arkov models.
\newblock In {\em 2009 IEEE International conference on multimedia and expo},
  pages 1218--1221. IEEE, 2009.

\bibitem{IHHMM}
Katherine Heller, Yee~Whye Teh, and Dilan Gorur.
\newblock Infinite hierarchical hidden {M}arkov {M}odels.
\newblock In David van Dyk and Max Welling, editors, {\em Proceedings of the
  Twelth International Conference on Artificial Intelligence and Statistics},
  volume~5 of {\em Proceedings of Machine Learning Research}, pages 224--231,
  Hilton Clearwater Beach Resort, Clearwater Beach, Florida USA, 16--18 Apr
  2009. PMLR.

\bibitem{CVDHMM}
S.~E. Levinson.
\newblock Continuously variable duration hidden {M}arkov models for automatic
  speech recognition.
\newblock {\em Computer Speech and Language}, 1:29--45, 1986.

\bibitem{HHMManimal}
Timo Adam, Christopher~A Griffiths, Vianey Leos-Barajas, Emily~N Meese,
  Christopher~G Lowe, Paul~G Blackwell, David Righton, and Roland Langrock.
\newblock Joint modelling of multi-scale animal movement data using
  hierarchical hidden {M}arkov models.
\newblock {\em Methods in Ecology and evolution}, 10(9):1536--1550, 2019.

\bibitem{HHMMmovement}
Nam~Thanh Nguyen, Dinh~Q Phung, Svetha Venkatesh, and Hung Bui.
\newblock Learning and detecting activities from movement trajectories using
  the hierarchical hidden {M}arkov model.
\newblock In {\em 2005 IEEE Computer Society Conference on Computer Vision and
  Pattern Recognition (CVPR'05)}, volume~2, pages 955--960. IEEE, 2005.

\bibitem{HHMMdementia}
Svebor Karaman, Jenny Benois-Pineau, Vladislavs Dovgalecs, R{\'e}mi M{\'e}gret,
  Julien Pinquier, R{\'e}gine Andr{\'e}-Obrecht, Yann Ga{\"e}stel, and
  Jean-Fran{\c{c}}ois Dartigues.
\newblock Hierarchical hidden {M}arkov {M}odel in detecting activities of daily
  living in wearable videos for studies of dementia.
\newblock {\em Multimedia tools and applications}, 69:743--771, 2014.

\bibitem{HHMMonline}
Parviz Asghari, Elnaz Soleimani, and Ehsan Nazerfard.
\newblock Online human activity recognition employing hierarchical hidden
  {M}arkov models.
\newblock {\em Journal of Ambient Intelligence and Humanized Computing},
  11:1141--1152, 2020.

\bibitem{Jain2016}
Miten Jain, Hugh~E. Olsen, Benedict Paten, and Mark Akeson.
\newblock The {O}xford {N}anopore {M}in{ION}: {D}elivery of nanopore sequencing
  to the genomics community.
\newblock {\em Genome Biology}, 17, 12 2016.

\bibitem{Lokatt}
Xuechun Xu, Nayanika Bhalla, Patrik St{\r a}hl, and Joakim Jald{\'e}n.
\newblock Lokatt: {A} hybrid {DNA} nanopore basecaller with an explicit
  duration hidden {M}arkov model and a residual {LSTM} network.
\newblock {\em bioRxiv}, 2022.

\bibitem{HMMRATAC}
Evan~D. Tarbell and Tao Liu.
\newblock Hmmratac: a hidden {M}arkov {M}odele{R} for {A}{T}{A}{C}-seq.
\newblock {\em Nucleic acids research}, 47:e91, 9 2019.

\bibitem{HHMM}
Shai Fine, Yoram Singer, and Naftali Tishby.
\newblock The hierarchical hidden {M}arkov {M}odel: {A}nalysis and
  {A}pplications.
\newblock {\em Machine Learning}, 32, 1998.

\bibitem{LTHHMM}
Kevin~P Murphy and Mark Paskin.
\newblock Linear-time inference in hierarchical {HMM}s.
\newblock In {\em Advances in Neural Information Processing Systems},
  volume~14. MIT Press, 2001.

\bibitem{durationviterbi}
Churbanov Alexander, Baribault Carl, and Winters-Hilt Stephen.
\newblock Duration learning for analysis of nanopore ionic current blockades.
\newblock {\em BMC Bioinformatics}, 8, 11 2007.

\bibitem{bioinfomatics}
Bro{\v{n}}a Brejov{\'a}, Daniel~G. Brown, and Tom{\'a}{\v{s}} Vina{\v{r}}.
\newblock The most probable labeling problem in {HMM}s and its application to
  bioinformatics.
\newblock In {\em Algorithms in Bioinformatics}, pages 426--437, Berlin,
  Heidelberg, 2004. Springer Berlin Heidelberg.

\bibitem{Chiron}
Haotian Teng, Minh~Duc Cao, Michael~B. Hall, Tania Duarte, Sheng Wang, and
  Lachlan~J.M. Coin.
\newblock Chiron: translating nanopore raw signal directly into nucleotide
  sequence using deep learning.
\newblock {\em GigaScience}, 2018.

\bibitem{Bonito}
Oxford Nanopore~Technologies plc.
\newblock Nanoporetech/{B}onito: A {P}ytorch basecaller for {O}xford {N}anopore
  reads.
\newblock \url{https://github.com/nanoporetech/bonito}, February 2020.

\bibitem{Hayashi2013}
Akira Hayashi, Kazunori Iwata, and Nobuo Suematsu.
\newblock Marginalized {V}iterbi algorithm for hierarchical hidden {M}arkov
  models.
\newblock {\em Pattern Recognition}, 46:3452--3459, 12 2013.

\bibitem{CNN-HMM}
Qiang Guo, Fenglei Wang, Jun Lei, Dan Tu, and Guohui Li.
\newblock Convolutional feature learning and hybrid {CNN}-{HMM} for scene
  number recognition.
\newblock {\em Neurocomputing}, 184:78--90, 2016.
\newblock RoLoD: Robust Local Descriptors for Computer Vision 2014.

\bibitem{DNN-HMM}
Qiujia Li, Chao Zhang, and Philip~C. Woodland.
\newblock Combining hybrid {DNN}-{HMM} {ASR} systems with attention-based
  models using lattice rescoring.
\newblock {\em Speech Communication}, 147:12--21, 2023.

\bibitem{BS}
Defense Technical~Information Center.
\newblock {DTIC} {ADA}049288: {S}peech understanding systems. {S}ummary of
  results of the five-year research effort at {C}arnegie-{M}ellon {U}niversity.
\newblock 1977.

\bibitem{improveBS}
Volker Steinbiss, Bach-Hiep Tran, and Hermann Ney.
\newblock Improvements in beam search.
\newblock In {\em Third international conference on spoken language
  processing}, 1994.

\bibitem{viterbiBS}
Monalisa Mazumdar, Mun-Ho Jeong, and Bum-Jae You.
\newblock An online optimal path decoder for {HMM} towards connected hand
  gesture recognition.
\newblock {\em IFAC Proceedings Volumes}, 41(2):736--741, 2008.
\newblock 17th IFAC World Congress.

\bibitem{BScrf}
Heng Zhang, Xiang-Dong Zhou, and Cheng-Lin Liu.
\newblock Keyword spotting in handwritten chinese documents using semi-{M}arkov
  conditional random fields.
\newblock {\em Engineering Applications of Artificial Intelligence}, 58:49--61,
  2017.

\bibitem{BSnayesiannetwork}
Xiaoyuan Zhu and Changhe Yuan.
\newblock Hierarchical beam search for solving most relevant explanation in
  {B}ayesian networks.
\newblock {\em Journal of Applied Logic}, 22:3--13, 2017.
\newblock SI:Uncertain Reasoning.

\bibitem{bitonicsorter}
K.~E. Batcher.
\newblock Sorting networks and their applications.
\newblock In {\em Proceedings of the April 30--May 2, 1968, Spring Joint
  Computer Conference}, AFIPS '68 (Spring), page 307–314, New York, NY, USA,
  1968. Association for Computing Machinery.

\bibitem{EDHMM}
ShunZheng Yu and Hisashi Kobayashi.
\newblock Practical implementation of an efficient forward-backward algorithm
  for an explicit-duration hidden {M}arkov model.
\newblock {\em IEEE Transactions on Signal Processing}, 54(5):1947--1951, 2006.

\bibitem{durationmodelHMM}
Yi-Jian Wu, Hisashi Kawai, Jinfu Ni, and Ren-Hua Wang.
\newblock Discriminative training and explicit duration modeling for
  {HMM}-based automatic segmentation.
\newblock {\em Speech Communication}, 47(4):397--410, 2005.

\bibitem{semiHMM}
Shun~Zheng Yu.
\newblock Hidden semi-{M}arkov models.
\newblock {\em Artificial Intelligence}, 174:215--243, 2 2010.

\bibitem{stringmatch}
Gonzalo Navarro.
\newblock A guided tour to approximate string matching.
\newblock {\em ACM Computing Surveys}, 33, 04 2000.

\bibitem{biopython}
Peter~JA Cock, Tiago Antao, Jeffrey~T Chang, Brad~A Chapman, Cymon~J Cox,
  Andrew Dalke, Iddo Friedberg, Thomas Hamelryck, Frank Kauff, Bartek
  Wilczynski, et~al.
\newblock Biopython: freely available {P}ython tools for computational
  molecular biology and bioinformatics.
\newblock {\em Bioinformatics}, 25(11):1422--1423, 2009.

\bibitem{cuda}
NVIDIA, Péter Vingelmann, and Frank~H.P. Fitzek.
\newblock Cuda, release: 10.2.89, 2020.

\bibitem{Minimap2}
Heng Li.
\newblock Minimap2: {P}airwise alignment for nucleotide sequences.
\newblock {\em Bioinformatics}, 34:3094–3100, 2018.

\bibitem{CTC}
Alex Graves, Santiago Fernández, Faustino Gomez, and Jürgen Schmidhuber.
\newblock Connectionist temporal classification: Labelling unsegmented sequence
  data with recurrent neural networks.
\newblock volume 2006, pages 369--376, 01 2006.

\end{thebibliography}
